\documentclass[letterpaper, 10 pt, conference]{ieeeconf}  

\IEEEoverridecommandlockouts                              

\usepackage{url}
\usepackage{graphics} 
\usepackage{epsfig} 
\usepackage{mathptmx} 
\usepackage{times} 
\usepackage{amsmath} 
\usepackage{amssymb}  
\usepackage{caption}
\usepackage{subfigure}
\usepackage{graphicx}
\usepackage{makecell}
\usepackage{float} 
\usepackage{color}
\usepackage{xcolor}
\usepackage{url}
\usepackage{hyperref}
\usepackage[nameinlink,noabbrev]{cleveref}
\usepackage[backend=bibtex]{biblatex}
\usepackage{multirow}
\usepackage{graphicx}
\usepackage{booktabs}

\title{\LARGE \bf
 BEVScope: Enhancing Self-Supervised Depth Estimation Leveraging Bird's-Eye-View in Dynamic Scenarios
}

\author{Yucheng Mao$^{1,2}$, Ruowen Zhao$^{1,3}$, Tianbao Zhang$^{1,4}$ and Hang Zhao$^{1*}$
\thanks{*Corresponding at: hangzhao@mail.tsinghua.edu.cn.}
\thanks{$^{1}$IIIS, Tsinghua University}%
\thanks{$^{2}$University of Science and Technology Beijing, 30 Xueyuan Road, Haidian District, Beijing, China. 
{\tt\small (maoyucheng@xs.ustb.edu.cn)}}%
\thanks{$^{3}$University of Chinese Academy of Sciences, 19 Yuquan Road, Shijingshan District, Beijing, China.
{\tt\small (zhaorewen20@mails.ucas.ac.cn)}}%
\thanks{$^{4}$Southeast University, No.2, Sipailou, Xuanwu District, Nanjing, China.
{\tt\small (tbzhangrobo@gmail.com)}}
}

\begin{document}

\maketitle
\thispagestyle{empty}
\pagestyle{empty}

\begin{abstract}
\noindent 
Depth estimation is a cornerstone of perception in autonomous driving and robotic systems. The considerable cost and relatively sparse data acquisition of LiDAR systems have led to the exploration of cost-effective alternatives, notably, self-supervised depth estimation. Nevertheless, current self-supervised depth estimation methods grapple with several limitations: (1) the failure to adequately leverage informative multi-camera views. (2) the limited capacity to handle dynamic objects effectively. To address these challenges, we present BEVScope, an innovative approach to self-supervised depth estimation that harnesses Bird's-Eye-View (BEV) features. Concurrently, we propose an adaptive loss function, specifically designed to mitigate the complexities associated with moving objects.  Empirical 
evaluations conducted on the Nuscenes dataset validate our
approach, demonstrating competitive performance. Code will be released at \href{https://github.com/myc634/BEVScope}{https://github.com/myc634/BEVScope}.

\end{abstract}



\begin{figure*}[!t]
  \centering
  \includegraphics[height=5cm,width=12cm]{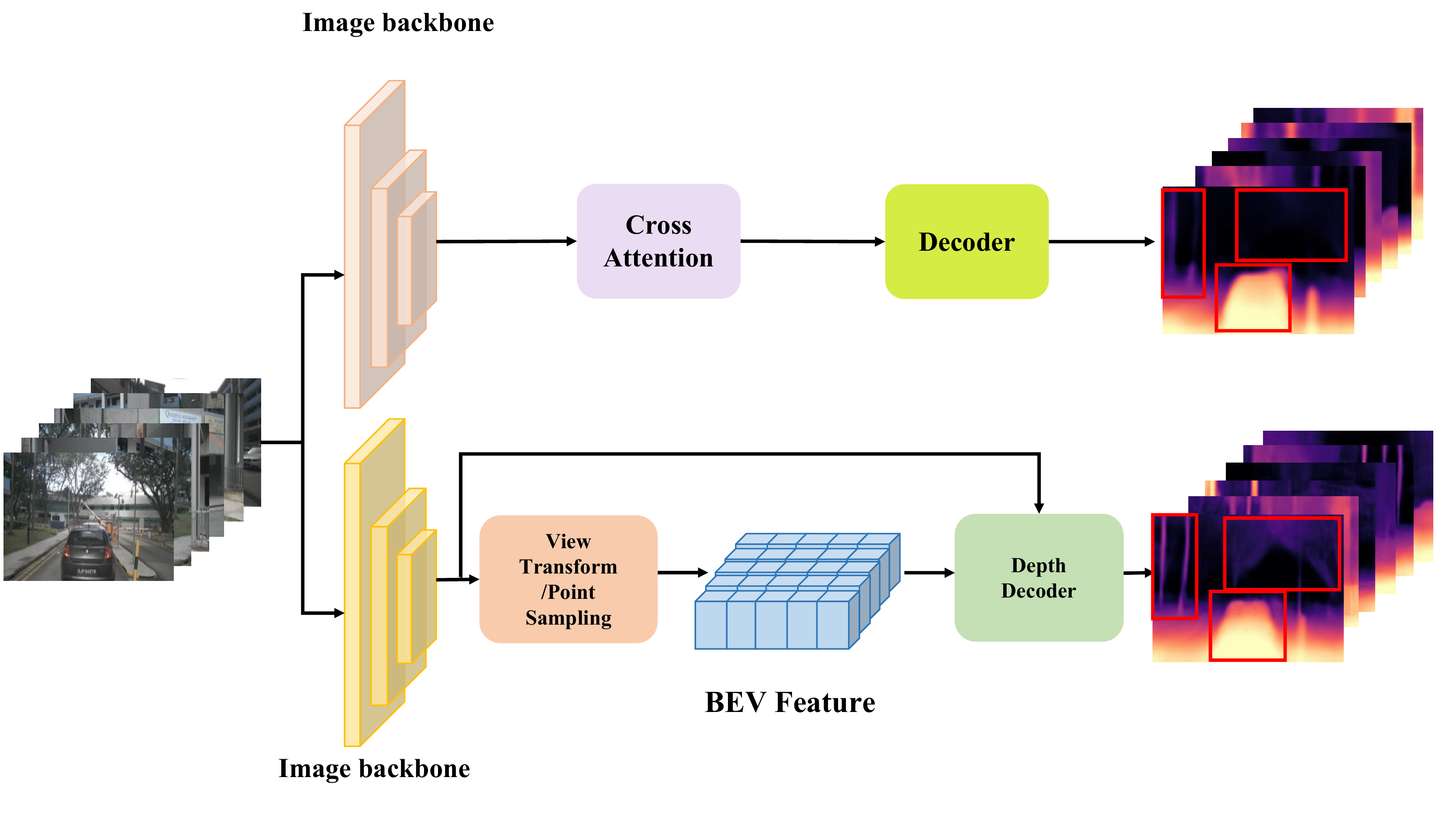}
  \caption{Comparison of BEVScope and camera-view-based depth estimation methods. The upper workflow illustrates camera-view-based depth estimation method \cite{Surrounddepth, MCDP, FSM}, which do not account for the geometric relationships between different perspective views. The lower workflow illustrates our proposed BEVScope method. \textbf{View Transform} signifies the BEV feature generation methods based on the LSS paradigm \cite{BEVDet, BEVDet4D, BEVDepth, LSS, SOLOFusion, BEVStereo}. Furthermore, \textbf{Point Sampling} refers to BEV feature generation methods that share similarities with \cite{BEVFromer, BEVFormerv2}}
  \label{fig1}
\end{figure*}

\section{Introduction}
\noindent In both the robotics and autonomous driving domains, the significance of 3D perception is paramount. Depth information, which acts as a vital link between 2D image inputs and the actual 3D environment, plays a crucial role \cite{LSS, monodepth, monodepth2}. However, the utilization of depth sensors, such as LiDAR, although effective, often faces obstacles owing to their significant cost and the relatively sparse nature of the data they provide. Conversely, cameras, despite lacking inherent depth information, present a cost-effective alternative that can capture an abundance of semantic information. Therefore, the challenge resides in the extraction of depth information from these 2D images, a task that depth estimation methodologies need to tackle effectively.

\noindent Owing to the significant cost associated with acquiring densely annotated depth maps, depth estimation often employs a self-supervised learning approach \cite{Surrounddepth, monodepth, zhou2017unsupervised}. The pioneering technique of self-supervised monocular depth estimation was introduced by \cite{zhou2017unsupervised}, which catalyzed a chain of subsequent advancements in this domain. The Full Surround Monodepth (FSM) \cite{FSM} methodology expanded on this technique, integrating a multi-view perspective for the first time. Additional enhancements were brought about by SurroundDepth \cite{Surrounddepth}, which augmented the cross-view interaction of information via the implementation of Structure-from-Motion (SfM), consequently facilitating the recovery of real-world scales. Multi-Camera Collaborative Depth Prediction (MCDP) \cite{MCDP} further propelled these advancements by introducing a depth consistency loss that refines depth information in regions captured by overlapping cameras. The prevalent methodology involves joint depth and pose prediction, which is utilized to map the target frame to the source frame, thereby computing photometric loss as the supervision signal. Nonetheless, this supervision signal may encounter challenges in addressing dynamic objects. However, they primarily concentrate on estimating depth from a camera view, resulting in a limited understanding of the geometric structure, which in turn hampers performance.

\noindent Significant advancements have been observed in autonomous driving tasks such as object detection and map segmentation, owing to the incorporation of Bird's-Eye-View (BEV) features \cite{DETR3D, BEVDet,BEVDet4D, BEVDepth,BEVFromer,BEVFormerv2,BEVStereo,LSS,SOLOFusion,lin2022sparse4d,PETR,Polarformer}. We advocate for the utilization of BEV features in fostering robust depth estimation methodologies. The proposed Bird's-Eye-View (BEV)-oriented depth estimation strategy supersedes the conventional camera-view-dependent methods by explicitly integrating critical geometric structures. Our BEV-based approach is specifically devised to facilitate superior extraction and integration of geometric attributes across varied image perspectives. We delve into diverse techniques that drive the interaction between BEV information and image data.

\noindent To overcome the complexities associated with rapidly moving objects in real-world scenarios, we further present an adaptive loss function. In situations where a substantial number of rapidly moving objects persist within proximate frames of a given scene, this could lead to the ineffectiveness of self-supervised photometric loss in overseeing these rapid elements. The proposed adaptive photometric loss function is designed to lessen the weight according to rapidly moving objects within the supervision signal. Consequently, this promotes more robust and precise depth estimations.

\noindent In summary, our contributions are threefold:
    
\textbf{(i)} We introduce a novel depth estimation method that harnesses the power of Bird's-Eye-View (BEV) features to integrate geometric cues, thereby enhancing self-supervised learning depth estimation. This proposal marks a first endeavor in the application of BEV features for depth map generation.

\textbf{(ii)} We introduce a novel loss function, tailored to address the complexities inherent in depth estimation for rapidly moving objects.

\textbf{(iii)} We evaluate our proposed method on popular multi-camera autonomous driving datasets and achieve a competitive performance compared to the current state-of-the-art methods.

\section{Related Work}

\label{gen_inst}

\subsection{BEV-based Visual Perception}
    
    \noindent Bird's Eye-View (BEV) in 3D perception has garnered considerable interest recently due to its capability to represent an entire scene by integrating multiple views of data captured from surrounding cameras. Contemporary methods for generating BEV features primarily fall into two categories: dense BEV-based methods \cite{BEVDet,BEVDet4D, BEVDepth,BEVFromer,BEVFormerv2,BEVStereo,LSS,SOLOFusion,Polarformer} and sparse query-based methods \cite{DETR3D,lin2022sparse4d,PETR,liu2022petrv2}.

    \noindent Dense BEV-based methods address transformation in a direct manner, with Lift-Splat-Shoot (LSS) \cite{LSS} as a prime example of this approach. BEVDepth \cite{BEVDepth} and BEVDet \cite{BEVDet} employ the LSS paradigm, proposing efficient frameworks for multi-camera 3D object detection from Bird's Eye-View. BEVFormer \cite{BEVFromer} is the first to incorporate sequential temporal modeling into multi-view 3D object detection, implementing temporal self-attention. Building on its prior version, BEVDet4D \cite{BEVDet4D} also exploits temporal cues from multi-camera images. Besides these methods that fuse short-term timestamp images to generate the BEV feature map, SOLOFusion \cite{SOLOFusion} utilizes multiple timesteps across a long-term history.

    \noindent Sparse query-based methods generally utilize learnable queries to aggregate 2D image features via an attention mechanism. DETR3D \cite{DETR3D} initially leverages object queries to predict 3D positions, subsequently back-projecting them to 2D coordinates to extract the corresponding features. PETR \cite{PETR}, on the other hand, encodes 3D position information into the image features, enabling direct querying with global 2D features.
\begin{figure*}[!t]
  \centering
  \includegraphics[height=8cm,width=13cm]{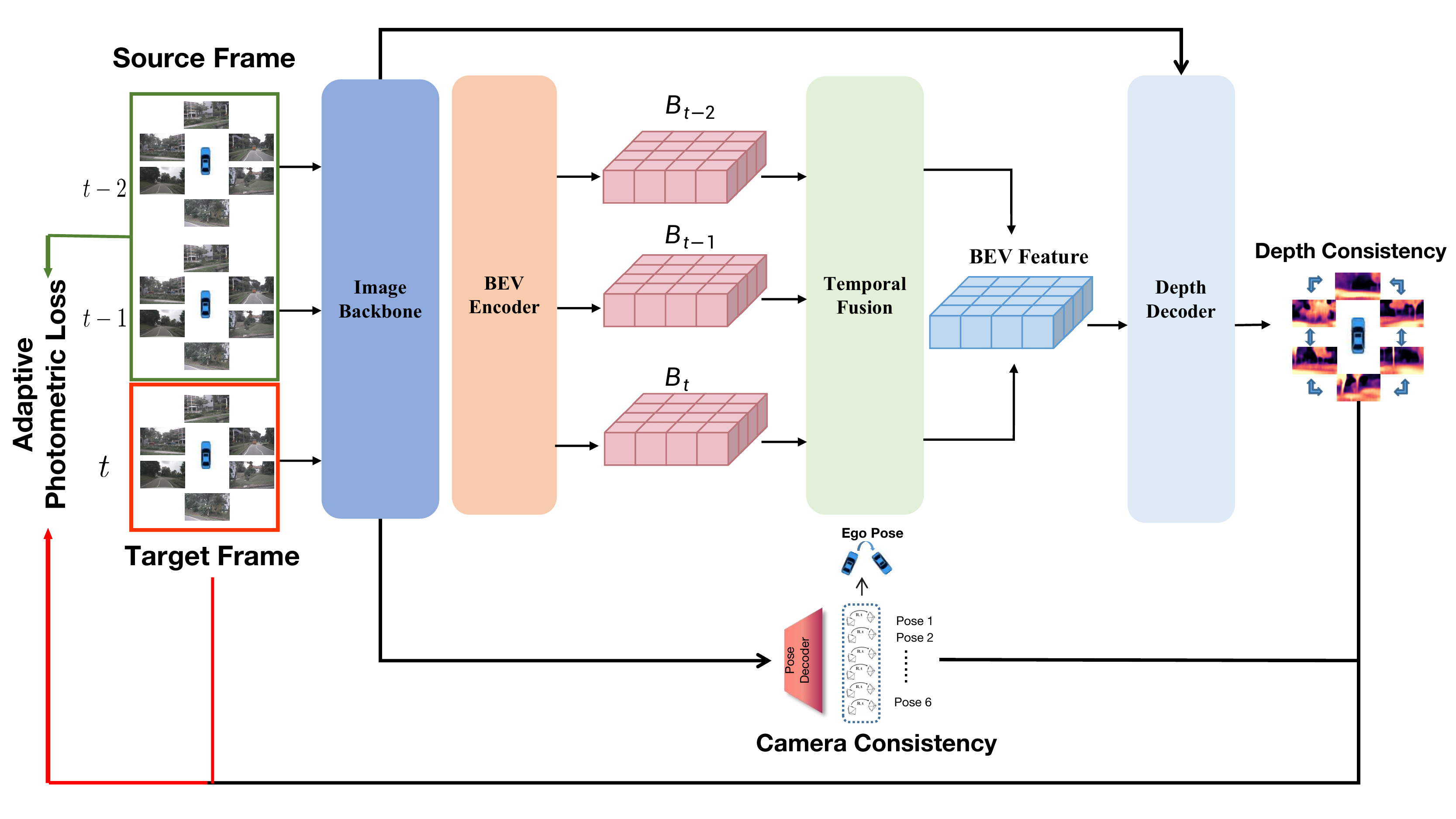}
  \caption{The Architecture of BEVScope: Our proposed BEVScope architecture encompasses a meticulously crafted depth estimation task head that facilitates the prediction of dense depth maps by leveraging both the existing image features and BEV features. To bolster the precision of depth estimation, particularly for fast-moving objects, we introduce an adaptive photometric loss function. Furthermore, for the temporal fusion of BEV features, we integrate a camera pose consistency loss function. The \textbf{BEV Encoder} component denotes the BEV generating methods employed, such as \cite{BEVDet, BEVDet4D, BEVFromer}. The \textbf{Temporal Fusion} model indicate as the temporal fusion block used in \cite{BEVDet4D, BEVFromer}.}
  \label{fig2}
\end{figure*}

\subsection{Self-Supervised Monocular Depth Estimation}

    \noindent Existing methodologies for self-supervised monocular depth estimation concurrently utilize depth and motion networks. These apply a photometric consistency loss, altered by the predicted depth and motion between target and source images \cite{zhou2017unsupervised,godard2019digging,bian2019unsupervised,zhao2020towards,yin2018geonet,wang2017orientation,zhou2019moving,ranjan2019competitive,chang2018pyramid}. Zhou et al. \cite{zhou2017unsupervised} were the pioneers in developing a self-supervised pipeline for the estimation of depth and ego-motion. Subsequent advancements by Godard et al. \cite{monodepth2} incorporated a minimal reprojection loss to cater to occlusions, a full-resolution multi-scale sampling methodology to minimize visual artifacts, and a self-masking loss to disregard outlier pixels. A more recent approach, the PackNet SfM by Guizilini et al. \cite{guizilini20203d}, integrates packing and unpacking blocks that exploit 3D convolutions to acquire dense appearance and geometric data in real-time. This technique was further enhanced by Poggi et al. \cite{poggi2020uncertainty}, who examined the estimation of uncertainty in this task and its effects on depth accuracy. Additionally, the Mono-Lite framework \cite{zhang2023lite} introduces a hybrid CNN and Transformer structure, yielding superior accuracy compared to Monodepth2 \cite{monodepth2} whilst reducing the model parameters. Despite the substantial advancements these works have brought to monocular depth estimation tasks, their extension to multi-view settings frequently encountered in autonomous driving scenarios is limited. Further, their capability to extract cross-view information effectively remains insufficient.

\subsection{Self-Supervised Surround-View Depth Estimation}

\noindent Recent studies have sought to address these challenges in multi-view settings. Guizilini et al. \cite{FSM} developed the Full Surround Monocular (FSM) approach, utilizing generalized spatio-temporal contexts and pose consistency constraints to facilitate self-supervised depth estimation with comprehensive surround multi-camera inputs. Similarly, the SurroundDepth methodology introduced by Wei et al. \cite{Surrounddepth} intertwines surrounding visual information across multiple scales via a cross-view transformer, while a joint pose estimation framework fully exploits the extrinsics in multi-camera settings. Further, Xu et al. \cite{MCDP} put forth the Multi-Camera Depth Prediction (MCDP) methodology, which introduces a depth consistency loss for minimizing disparities in depth for overlapping areas between depth maps estimated from different cameras. However, these methodologies predominantly rely on 2D feature extraction models, and as such, fail to fully leverage the powerful feature extraction capabilities of BEV-based visual perception frameworks.

\begin{table*}[htbp]
\centering
\caption{Quantitative results for depth estimation on the nuScenes \cite{nuscenes} dataset. All methods are trained and tested with the same experimental settings. \dag~indicates that during the training process, we replace the images of adjacent frames in the keyframe with the images from adjacent frames in the sweeps of the nuScenes \cite{nuscenes} dataset. $*$ indicates the implementation result in \cite{Surrounddepth}.}
\label{main_table}
\begin{tabular}{ccccccc}
\toprule
Methods & Abs Rel($\downarrow$) & Sq Rel($\downarrow$) & RMSE($\downarrow$) & a1($\uparrow$) & a2($\uparrow$) & a3($\uparrow$) \\
\midrule
FSM\cite{FSM} & 0.299 & - & - & - & - & - \\
FSM$*$\cite{FSM} & 0.334 & 2.845 & 7.786 & 0.508 & 0.761 & 0.894 \\
SurroundDepth\cite{Surrounddepth} & 0.245 & 3.067 & 6.835 & 0.719 & \textbf{0.878} & 0.935 \\
SurroundDepth\dag \cite{Surrounddepth} & 0.240 & 2.869 & 6.753 & 0.719 & 0.877 & 0.935 \\
MCDP\dag\cite{MCDP} & 0.237 & 3.030 & 6.822 & 0.719 & - & - \\
\midrule
BEVScope (w/o adaptive loss) & 0.239 & 2.777 & 6.842 & 0.711 & 0.872 & 0.933 \\
BEVScope\dag (w/o adaptive loss) & 0.236 & \textbf{2.122} & 6.884 & 0.678 & 0.862 & 0.930 \\
BEVScope\dag & \textbf{0.232} & 2.652 & \textbf{6.672} & \textbf{0.720} & 0.876 & \textbf{0.936} \\
\bottomrule
\end{tabular}
\end{table*}
\section{Method}

\subsection{Overview}
\noindent In this work, we put forth a Bird's-Eye-View (BEV)-based module designed for surround-view depth estimation. As depicted in Figure [\ref{fig2}], our approach comprises of the BEV feature fusion module. The following sections will detail the definition of the surround view depth estimation task and elaborate on the underlying principles of the BEVScope approach.

\subsection{Formulation} 

\noindent Considering two temporally sequential surround-view images, $\mathbf{I}_i^t$ and $\mathbf{I}_i^{t+1}$, procured by multiple cameras $\mathbf{C}_i$ (where $i \in {1, 2, ..., 6}$ represents the number of cameras) encompassing a comprehensive view around a vehicle, our approach is tasked with estimating the depth $\mathbf{D}_i^t$ and ego-motion $\mathbf{P}_i^{t \rightarrow s}$ of each camera $\mathbf{C_i}$. We make the assumption that the camera intrinsics $\mathbf{K}_i$ and extrinsics $\mathbf{E}_i$ of each viewpoint are known factors.

\noindent The depth network denoted as \textbf{\emph{F}} and the pose network represented by \textbf{\emph{G}} are trained by minimizing a per-pixel photometric reprojection loss in a self-supervised manner \cite{monodepth, zhou2017unsupervised}. The formulation of surround-views depth estimation can be expressed as follows:

\begin{align*}
D_i^{t} = F({I}_i^t) , \ \ \
P_i^{t \rightarrow s} = G({I}_i^t,{I}_i^s) \tag{1}
\end{align*}

\subsection{Depth Estimation Decoder}
\begin{figure}[h]
  \centering
  \includegraphics[width=8cm]{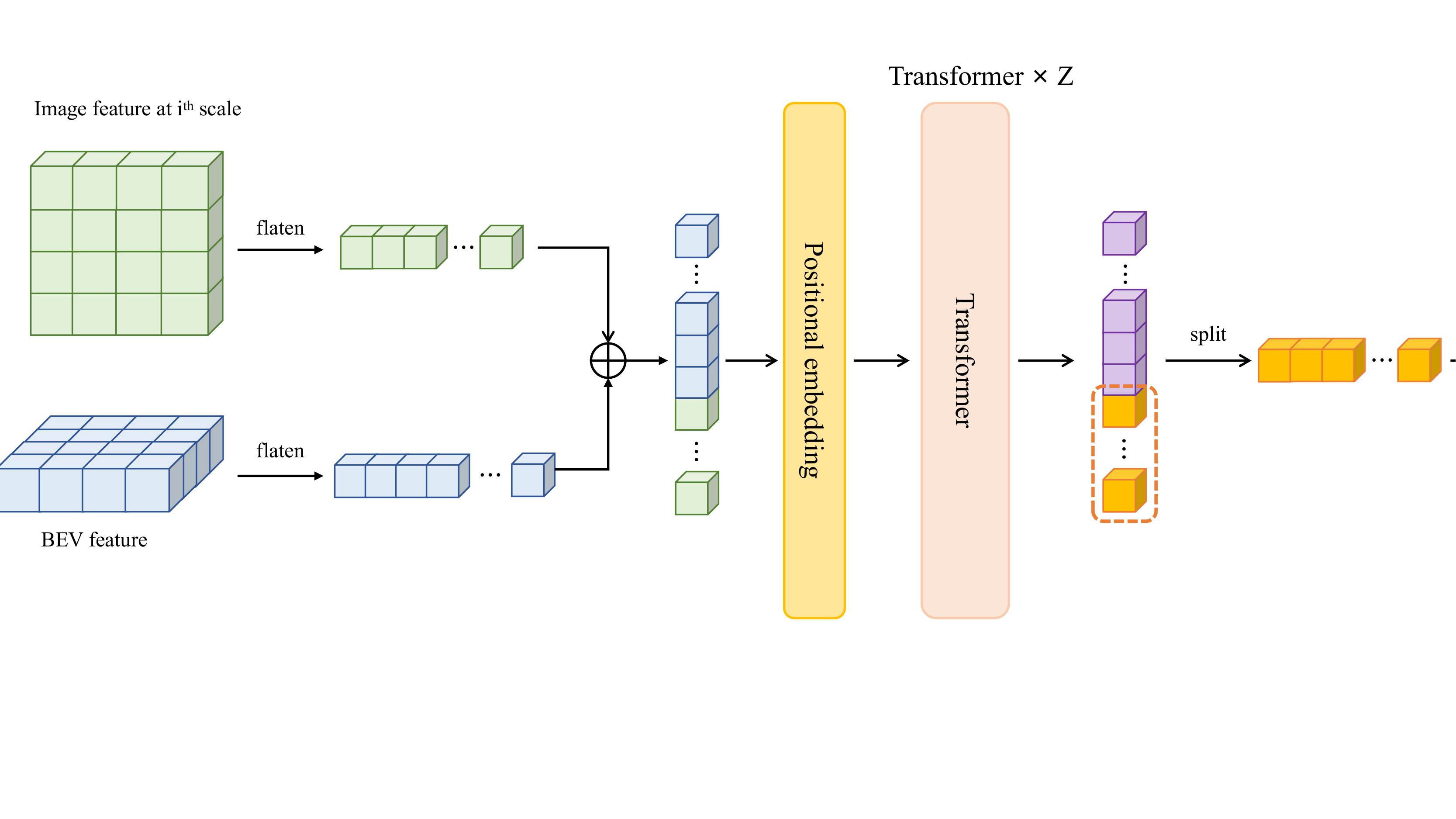}
  \caption{Depth Head of BEVScope. For image feature map in each $k$-scales, we use DS-Conv to downsample it and concat with the flattened BEV feature.Then, we interact the image feature and BEV feature through a self-attention mechanism.}
  \label{fig3}
\end{figure}

\noindent The architecture of our proposed depth estimation component within BEVScope is demonstrated in Figure[\ref{fig3}]. \\
\noindent \textbf{Network parameterization}  Our proposed \textbf{Scope Head} is to predict the depth map in
\textbf{k}-scales from \textbf{k}-scales feature maps and the BEV feature. We define $F_\mathrm{img}^{k} \in \mathbb{R}^{N \times H_\mathrm{img}^{k} \times W_\mathrm{img}^{k} \times C_\mathrm{img}^{k}} $ to represent the feature map, extracted from the surround $N$ cameras in \textbf{k}-scales from a weight-sharing image backbone. We denote $F_\mathrm{BEV} \in \mathbb{R}^{1 \times H_\mathrm{BEV} \times W_\mathrm{BEV} \times C_\mathrm{BEV}} $ as the BEV features generated from BEV encoder such as \cite{BEVDet, BEVDet4D, BEVFromer}.  In our principal experiment, we utilize BEVFormer \cite{BEVFromer} as the BEV encoder. Here, we use $T$ to represent our proposed depth estimation task head $(i \in k)$:
\begin{align*}
F_\mathrm{depth}^{i} = T(F_\mathrm{img}^{i}, F_\mathrm{BEV}) \tag{2}
\end{align*}

\noindent \textbf{Image-BEV Feature Fusion}  Inspired by Vision Transformers \cite{ViT}, we transform the BEV features into patches. This transformation is based on the observation that feature maps from various perspective views display pronounced correlations with distinct regions in the BEV features. Specifically, we reshape the BEV feature $F_\mathrm{BEV} \in \mathbb{R}^{1 \times H_\mathrm{BEV} \times W_\mathrm{BEV} \times C_\mathrm{BEV}} $ into a sequence of flattened 2D patches $F_\mathrm{BEV} \in \mathbb{R}^{1 \times N_\mathrm{BEV} \times C_\mathrm{BEV}} $ where $(H_\mathrm{BEV}, W_\mathrm{BEV})$ defines the shape of the original BEV feature, $(P_\mathrm{BEV}, P_\mathrm{BEV})$ determines the size of each BEV patch, and $N = (H_\mathrm{BEV} \times W_\mathrm{BEV}) / P_\mathrm{BEV}^{2}$ is the resulting quantity of patches.

 \noindent Additionally, drawing on the Vision Transformer \cite{ViT} approach, we consider the image feature as a token symbolizing the BEV feature, enabling effective combination of the two types of features. To mitigate computational cost associated with the quadratic term, following \cite{Surrounddepth}, we employ a depthwise separable convolution (DS-Conv) \cite{chollet2017xception, howard2017mobilenets} prior to image-BEV fusion layers to downsample the extensive feature maps into lower-resolution ones with same channel numbers, denoted as $F_\mathrm{img}^{k} \in \mathbb{R}^{N \times h_\mathrm{img}^{k} \times w_\mathrm{img}^{k} \times C_\mathrm{img}^{k}} $. We flatten the image feature from varying perspective views $F_\mathrm{img}^{k} \in \mathbb{R}^{N \times (h_\mathrm{img}^{k} \times w_\mathrm{img}^{k}) \times C_\mathrm{img}^{k}} $ and concatenate the flattened image feature map with the BEV feature. The feature in each scale is formulated as follows, for $i\in k$:

\begin{align*}
F_\mathrm{all}^{i} = Concat(F_\mathrm{img}^{i}, F_\mathrm{BEV}) \tag{3}
\end{align*}

 \noindent We then construct $Z$ Image-BEV fusion self-attention layers with the objective of effectively integrating the two types of features. The image-BEV fusion layer is formulated as $(i \in k)$:

\begin{align*}
F_\mathrm{all}^{i} = \mathrm{Softmax}((F_\mathrm{all}^{i})^{T}F_\mathrm{all}^{i})/\sqrt{d_{F_\mathrm{all}^{i}}})F_\mathrm{all}^{i} \tag{4}
\end{align*}

\noindent Upon completing the Image-BEV fusion layers, to restore the depth map from the fused feature, we split the $F_\mathrm{all}^{i}$ into two sections, each corresponding to the original shape of $F_\mathrm{img}^{i}$ and $F_\mathrm{BEV}$. We disregard the $F_\mathrm{BEV}$ and employ $F_\mathrm{img}^{i}$ to upsample to the original shape and predict the dense depth map.

 \noindent While projecting the features from Bird's-Eye-View (BEV) into different perspectives is a straightforward approach, it often results in sparse features that hinder effective upsampling and accurate depth learning. To overcome this limitation, we propose a specially designed depth estimation task head to address this drawback.

\subsection{Adaptive Photometric Loss Function for Fast Moving Objects}

 \noindent This section introduces a loss function with adaptive perception to ameliorate this issue.

\noindent \textbf{Traditional Photometric Loss }  Traditionally, the supervision signal in a self-supervised depth estimation task is formulated as follows: Given a single image $I_{t}$ input at the $t$ frame, the system predicts its corresponding depth map $D_{t}$. Concurrently, the pose network leverages temporally adjacent images to estimate the relative pose $T_{t \rightarrow t+n}$ between the target image $I_{t}$ and the source image $I_{t+n}$. Based on the estimated depth map $D_{t}$, relative pose $T_{t \rightarrow t+n}$, and camera intrinsics matrix $K$, we can warp the image feature from the $t$ frame to the $t+n$ frame, denoted as $I_{t \rightarrow t+n}$. The system then minimizes the per-pixel minimum photometric re-projection error \cite{monodepth,zhou2017unsupervised} $L_{p}$ as:
\begin{align*}
L_p=\min pe\left(I_t, I_{t+n \rightarrow t}\right) \tag{5}
\end{align*}
where $pe(\centerdot)$ represents the photometric error, which consists of the $L_{1}$ error and the Structural Similarity (SSIM):
\begin{align*}
p e\left(I_a, I_b\right)= & \alpha\left(1-\operatorname{SSIM}\left(I_a, I_b\right)\right) + (1-\alpha)\left|I_a-I_b\right|_1 \tag{6}
\end{align*}

 \noindent However, this supervision signal struggles with handling rapidly moving objects. Such objects may not maintain the same position in adjacent frames. 
 
\noindent \textbf{Adaptive Photometric Loss Function}  Hence, an adaptive loss function is proposed to tackle this problem. Specifically, we calculate the per-pixel SSIM value of adjacent frames ${{I}_i^t,{I}_i^s}$. This computation yields matrix $S\in \mathbb{R}^{H \times W}$, which is utilized as a weight when computing the photometric loss. The design intent here is to increase model focus on areas with smaller SSIM values between adjacent frames. Therefore, the difference between the identity matrix and the SSIM matrix $(I-S) \in \mathbb{R}^{H \times W}$ is employed as the function weight. This approach effectively directs the network to pay less attention to areas with significant frame transitions. As a result, the adaptive loss function is reformulated as:
\begin{align*}
L_p^{ada} = (I-S)L_{p} \tag{7}
\end{align*}

 \noindent In the equation above, $L_{p}$ denotes the original photometric loss.

\begin{figure}[htbp]
    \centering
    \subfigure[Original Image]
    {
        \begin{minipage}[b]{.49\linewidth}
            \centering
            \includegraphics[height=3.1cm,width=4.2cm]{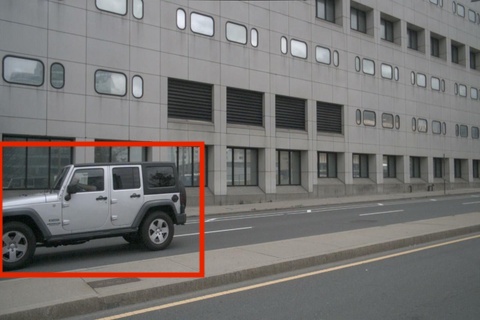}
        \end{minipage}
        \begin{minipage}[b]{.49\linewidth}
            \centering
            \includegraphics[height=3.1cm,width=4.2cm]{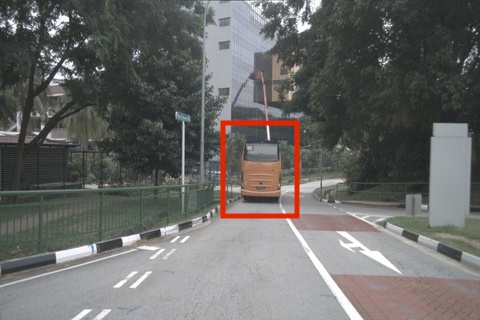}
        \end{minipage}}
    \qquad    
    \subfigure[BEVScope(w/o adaptive mask)]
    {
     	\begin{minipage}[b]{.49\linewidth}
            \centering
            \includegraphics[height=3.1cm,width=4.2cm]{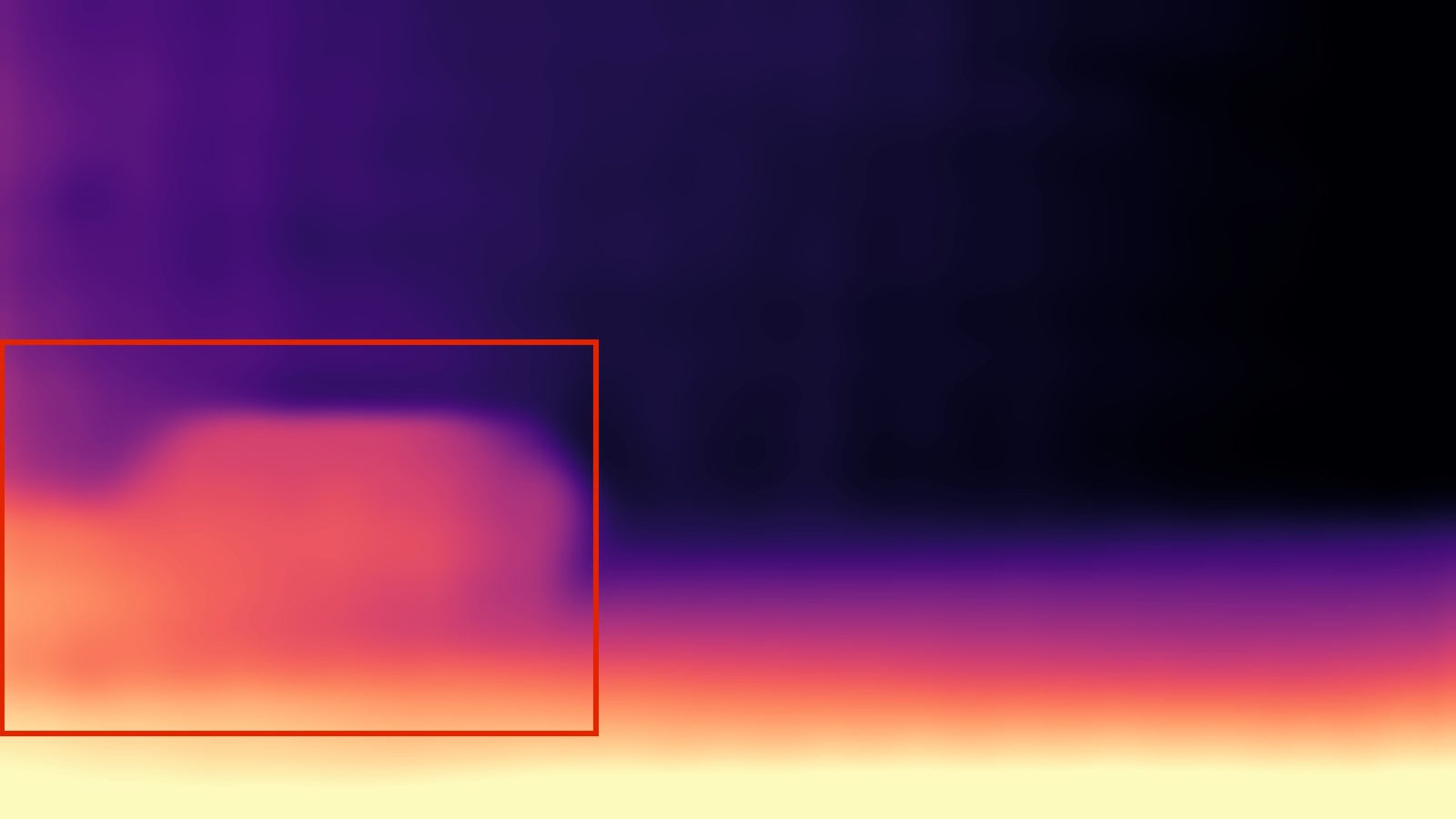}
        \end{minipage}
        \begin{minipage}[b]{.49\linewidth}
            \centering
            \includegraphics[height=3.1cm,width=4.2cm]{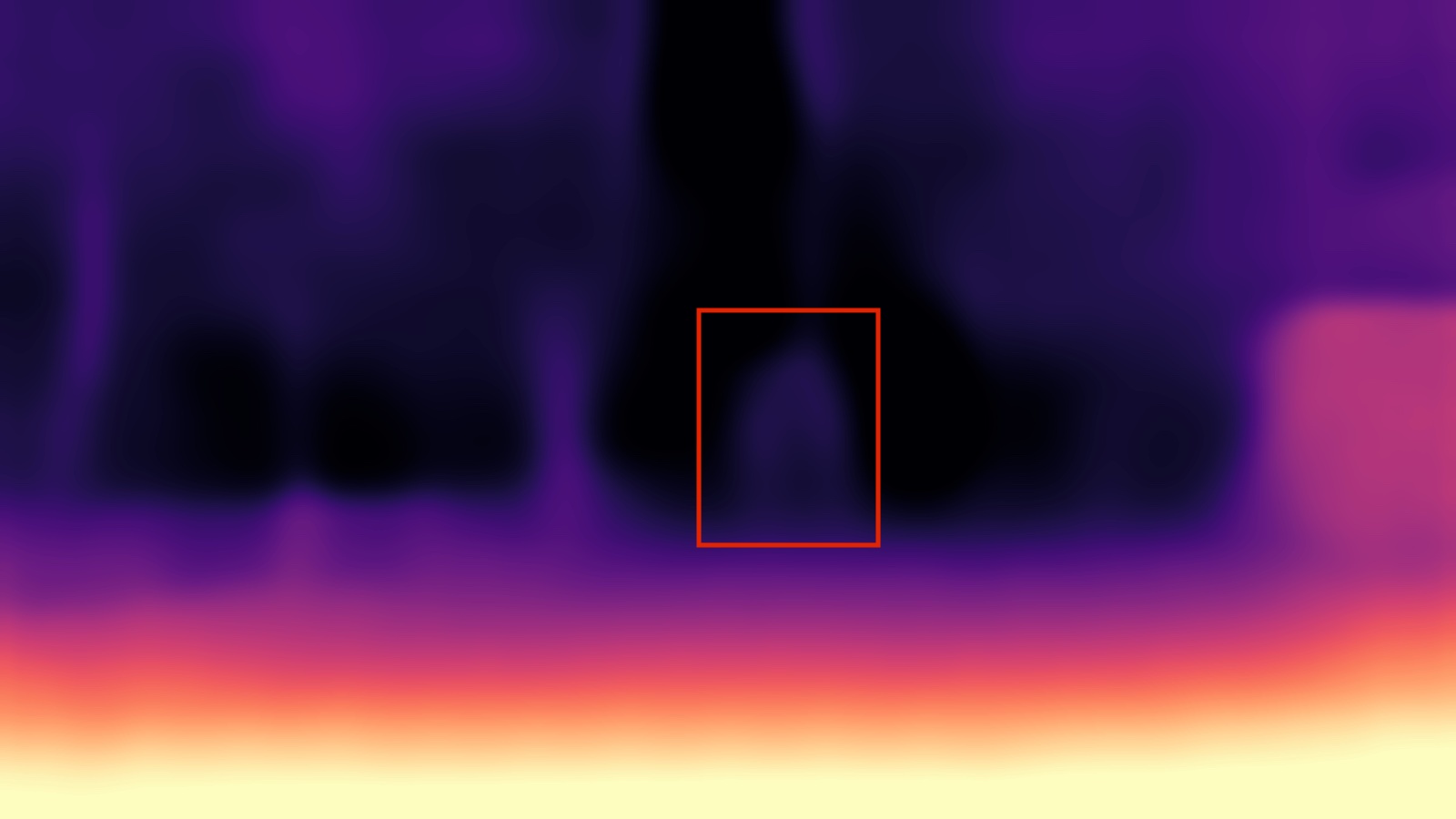}
        \end{minipage}}
    
    \subfigure[BEVScope(w adaptive mask)]
    {
    \begin{minipage}[b]{.49\linewidth}
        \centering
        \includegraphics[height=3.1cm,width=4.2cm]{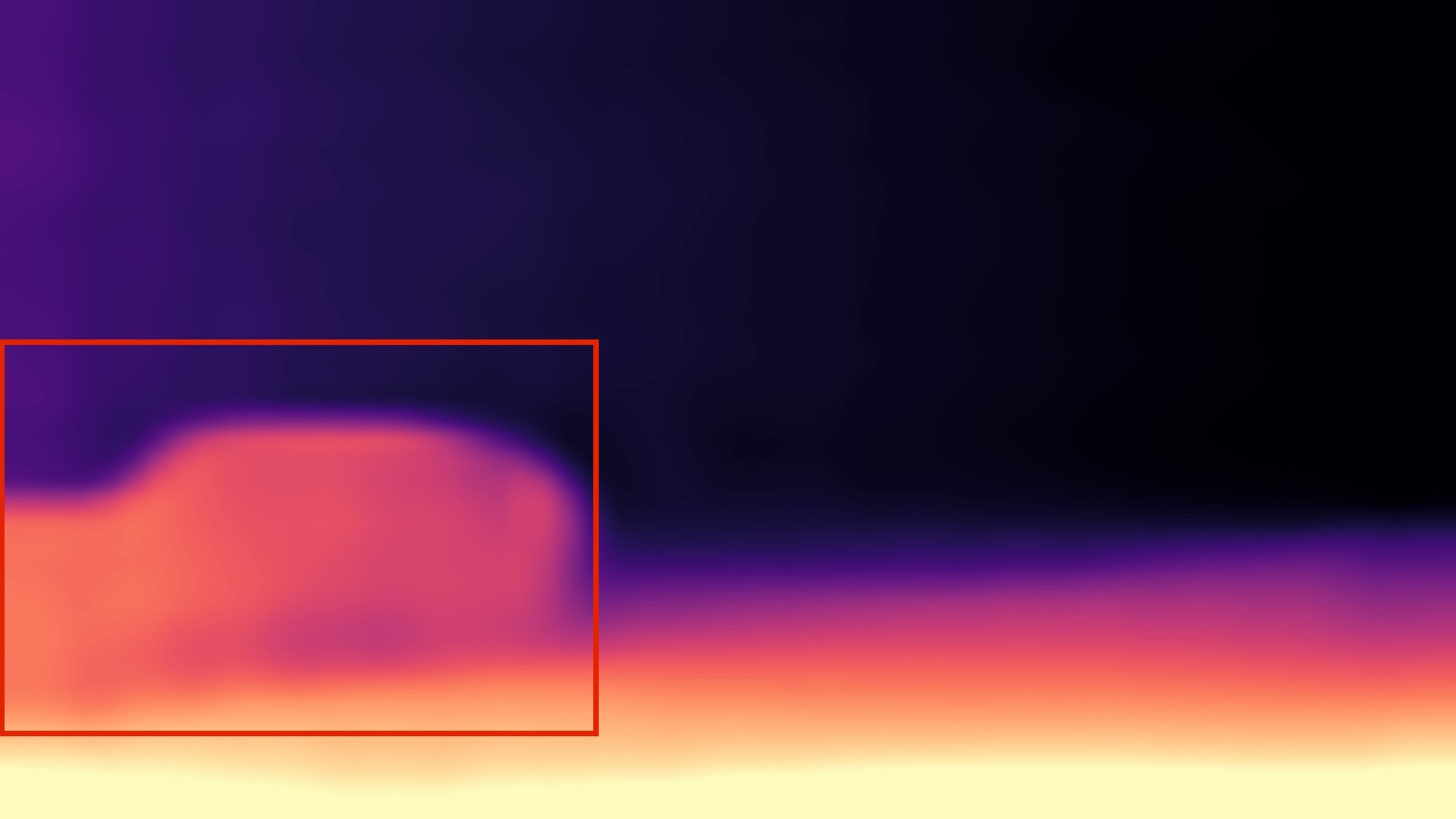}
    \end{minipage}
    \begin{minipage}[b]{.49\linewidth}
        \centering
        \includegraphics[height=3.1cm,width=4.2cm]{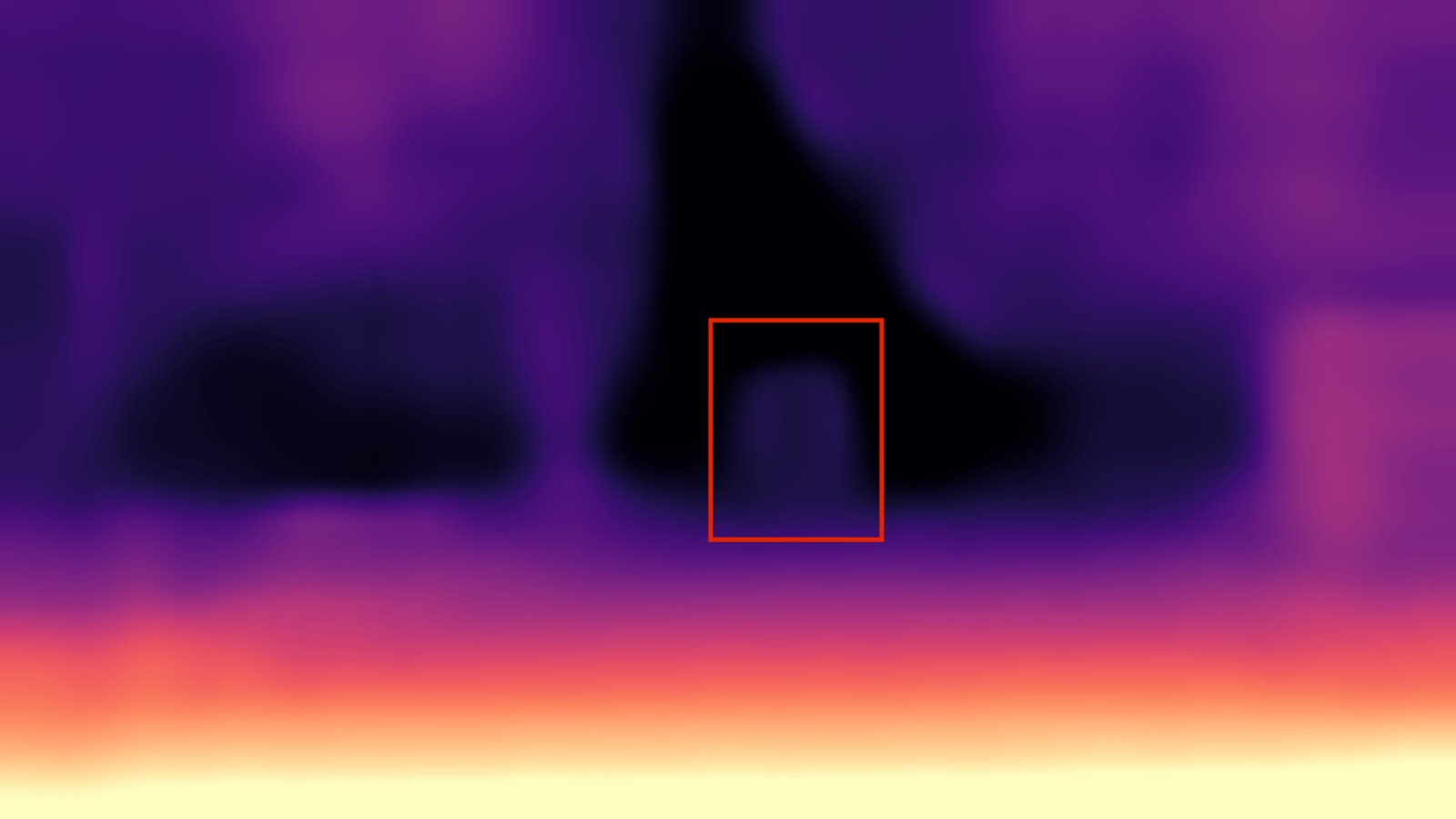}
    \end{minipage}}
    \caption{Demonstration of Adaptive Loss Function Effectiveness. We infer from the original image on BEVScope, comparing the results of two different scenarios: one using an adaptive loss function during training, and the other not. As illustrated by the contrast between Fig. (b) and Fig.(c), it is evident that the adaptive loss function improves the depth estimation for rapidly moving objects.}
    \label{fig4}
\end{figure}

\subsection{Incorporating Camera Pose Consistency}
 \noindent In observing prior methodologies \cite{FSM}, it was found that there was a lack of sufficient coupling between camera pose consistency and the task of depth estimation. This led to a lack of comprehensive attention to, and integration of, these two crucial aspects. To tackle this issue, we introduce an additional loss function, namely, a camera pose consistency constraint loss, which facilitates effective alignment of temporal BEV features. The benefit of this additional loss function is that it strengthens the relationship between camera poses and depth estimation, thus enhancing the overall performance of the system.

 \noindent The camera features are input to the pose decoder, which then predicts the pose for each camera, designated as $P_i^{camera}$ where $i \in {1, 2, ..., 6}$.

 \noindent The predicted camera poses at this stage are all situated in the coordinate system of their respective views. To impose an explicit constraint on the consistency across surround views, we transform the predicted camera poses into the ego coordinate system, $P_i^{ego}$, where $i \in {1, 2, ..., 6}$. Subsequently, we constrain the l1 norm of the vehicle's ego pose, $P_i^{ego}$.

\begin{align*}
L_{pose} = \Vert P_i^{ego} \Vert_1 , \ \ i \in {1, 2, ..., 6} \tag{8}
\end{align*}

 \noindent The final loss function for our model is then given by:

\begin{align*}
L_{final} = L_{p}^{ada} + L_{pose} + L_{depth} \tag{9}
\end{align*}

 \noindent In the above, $L_{p}^{ada}$ refers to the adaptive photometric loss function, $L_{pose}$ corresponds to the proposed ego pose loss, and $L_{depth}$ denotes the depth consistency loss, as described in \cite{MCDP}.

\section{Experiments}

 \noindent This section provides a comparative analysis of our proposed methodology with existing state-of-the-art self-supervised surround-view depth estimation techniques [cite] on the nuScenes \cite{nuscenes} datasets. Additionally, an ablation study is conducted to ascertain the performance of various aspects of our proposed method. Visual results of fast-moving objects as predicted by our BEVScope and other leading depth estimation methods are also presented for clarity.

\subsection{Experimental Procedure}
 \noindent Our experiments employ depth and pose networks that share the same backbone structure as the Monodepth2 model \cite{monodepth2}. Specifically, we utilize the ResNet34 \cite{he2016deep} with ImageNet \cite{russakovsky2015imagenet} pretrained weights as the encoder for all experiments in line with the SurroundDepth approach \cite{Surrounddepth}, inclusive of baseline methods. The depth maps were refactored based on varying focal lengths of the surrounding cameras, as detailed in \cite{lee2019big}. We incorporated Z = 8 image-BEV fusion layers for each scale, with all features downsampled to a 15x10 size prior to BEV fusion across both datasets. For a fair comparison with SurroundDepth and MCDP techniques, we used the same sweeps data from the nuScenes dataset as employed in the MCDP method \cite{MCDP}.


\subsection{Ablation Study}

 \noindent This subsection presents the results of the ablation studies conducted to validate the effectiveness of individual modules within our proposed framework. All experiments were executed on the nuScenes dataset.

\noindent\textbf{Evaluation of BEV Feature Generation Techniques:} We conducted a series of experiments employing diverse BEV generation techniques \cite{BEVDet, BEVDet4D, BEVFromer} to evaluate the adaptability of our proposed task head. Table \ref{table2} presents the performance of our task head using various BEV feature generation methods.

\begin{table}[h]
\setlength{\abovecaptionskip}{0pt}
\setlength{\belowcaptionskip}{-0cm}
\caption{Ablation Study  of Depth Estimation with Different BEV Feature Generation Techniques}
\label{table2}
\begin{center}
\begin{tabular}{|c|c|c|c|c|}
\hline
Methods& Abs Rel($\downarrow$) & Sq Rel($\downarrow$) & RMSE($\downarrow$) & a1($\uparrow$) \\
\hline
BevDet \cite{BEVDet} & 0.250 & 2.876 & 7.119 & 0.689 \\
\hline
BevDet4D \cite{BEVDet4D} & 0.248 & 2.682 & 7.070 & 0.713
\\
\hline
BEVFromer \cite{BEVFromer} & \textbf{0.232} & \textbf{2.652} & \textbf{6.672} & \textbf{0.720} \\
\hline
\end{tabular}
\end{center}
\end{table}
\noindent\textbf{Adaptive Loss Function:} Fig. \ref{fig4} visually demonstrates the effectiveness of the adaptive loss function in handling fast-moving objects. Additionally, Table \ref{table3} evaluates our model with and without the adaptive loss function on scenes containing fast-moving objects, highlighting the significance of this approach. Notably, we have sleeted the scenes which contain more than one fast-moving objects in the nuScenes \cite{nuscenes} validation set for evaluation.

\begin{table}[h]
\setlength{\abovecaptionskip}{0pt}
\setlength{\belowcaptionskip}{-0cm}
\caption{Ablation Study For Adaptive Loss Function}
\label{table3}
\begin{center}
\begin{tabular}{|c|c|c|c|c|c|c|}
\hline
Methods& Abs Rel($\downarrow$) & Sq Rel($\downarrow$) & RMSE($\downarrow$) & a1($\uparrow$) \\
\hline
\makecell{w/o adaptive loss} & 0.230 & \textbf{2.142} & 7.178 & 0.684 \\
\hline
with adaptive loss & \textbf{0.226} & 2.685 & \textbf{6.985} & \textbf{0.728} \\
\hline
\end{tabular}
\end{center}
\end{table}

\noindent\textbf{Camera Consistency Loss Function:}Table \ref{table4} shows the effectiveness of the camera consistency loss function in aligning the BEV feature in the temporal BEV generating method.

\begin{table}[h]
\setlength{\abovecaptionskip}{0pt}
\setlength{\belowcaptionskip}{-0cm}
\caption{Ablation Study For Camera Consistency Loss Function (cam consis. refer to camera consistency)}
\label{table4}
\begin{center}
\begin{tabular}{|c|c|c|c|c|c|c|}
\hline
Methods& Abs Rel($\downarrow$) & Sq Rel($\downarrow$) & RMSE($\downarrow$) & a1($\uparrow$) \\
\hline
\makecell{w/o cam consis.}& 0.246 & 3.696 & 7.273 & 0.729 \\
\hline
with cam consis. & \textbf{0.232} & \textbf{2.652} & \textbf{6.672} & \textbf{0.720} \\
\hline
\end{tabular}
\end{center}
\end{table}

\noindent\textbf{Patch Embedding on BEV Feature:}Table \ref{table5} illustrate the importance of patch embedding

\begin{table}[h]
\caption{Ablation Study For Patch Embedding on BEV Feature}
\label{table5}
\begin{center}
\begin{tabular}{|c|c|c|c|c|c|c|}
\hline
Methods& Abs Rel($\downarrow$) & Sq Rel($\downarrow$) & RMSE($\downarrow$) & a1($\uparrow$) \\
\hline
\makecell{w/o patch embedding} & 0.248 & 3.46 & 7.192 & 0.727 \\
\hline
with patch embedding & \textbf{0.232} & \textbf{2.652} & \textbf{6.672} & \textbf{0.720} \\
\hline
\end{tabular}
\end{center}
\end{table}

\section{CONCLUSIONS}
    In this paper, we propose a self-supervised multi-camera depth estimation method named \textbf{BEVScope}. We leverage the advantages of BEV features in fusing multi-view information, while introducing new constraints to address the complexity of joint pose estimation and mutual consistency in multi view depth maps by fully utilizing the camera consistency and adaptive Loss Function. Our method achieves competitive performance on multi camera depth estimation datasets, such as NuScenes\cite{nuscenes} datasets.










\section*{References}




\printbibliography[heading=none]


\section{APPENDIX}

\subsection{Datasets}

    We conduct our experiments on NuScenes Dataset \cite{nuscenes}, which is composed of 1000 sequences of outdoor scenes in Boston and Singapore and each sequence has an approximate duration of 20 seconds. The dataset is officially partitioned into training, validation, and testing subsets, comprising 700, 150, and 150 sequences respectively. In each sample, we have access to the six surrounding cameras along with their corresponding calibrations and neighbour frame information \cite{Surrounddepth}. 
    
\subsection{Implementation Details}
    
    We train all models with 5 epochs and a batch size of 1, where each batch contains 6 multi-view images, per NVIDIA RTX 3090 GPU. In line with the previous work, specially monodepth2\cite{monodepth2}, we set SSIM weight to 0.85 and L1 weight to 0.15 respectively in the reprojection loss. Furthermore, we assign a depth smooth weight of 1e-3 to ensure smoothness in the depth maps. In addition to these settings, we introduce our proposed constraints. We set both camera consistency and depth consistency loss weight as 1e-4 to enforce the consistency between the estimated camera poses and depth maps.

\begin{figure*}[t!]
    \centering
    \subfigure[Original Image]
    {
        \begin{minipage}[b]{.15\linewidth}
            \centering
            \includegraphics[height=1.5cm,width=3cm]{visual6/289_0_rgb.jpg}
        \end{minipage}
        \begin{minipage}[b]{.15\linewidth}
            \centering
            \includegraphics[height=1.5cm,width=3cm]{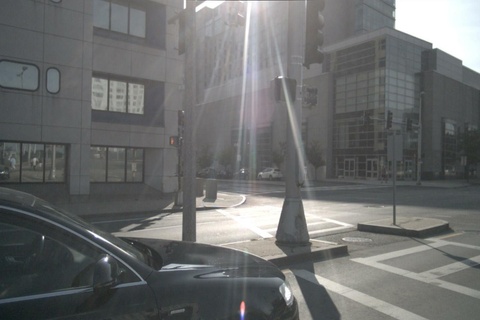}
        \end{minipage}
        \begin{minipage}[b]{.15\linewidth}
            \centering
            \includegraphics[height=1.5cm,width=3cm]{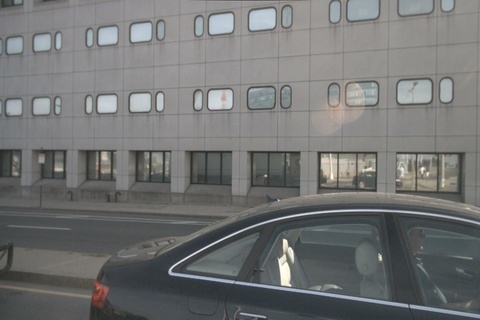}
        \end{minipage}
        \begin{minipage}[b]{.15\linewidth}
            \centering
            \includegraphics[height=1.5cm,width=3cm]{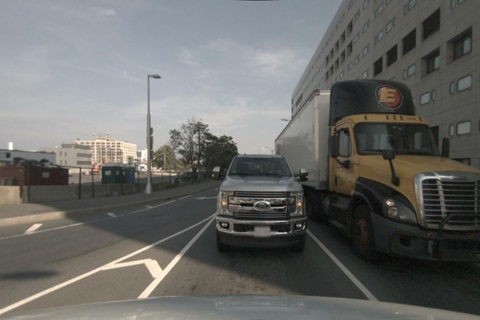}
        \end{minipage}
        \begin{minipage}[b]{.15\linewidth}
            \centering
            \includegraphics[height=1.5cm,width=3cm]{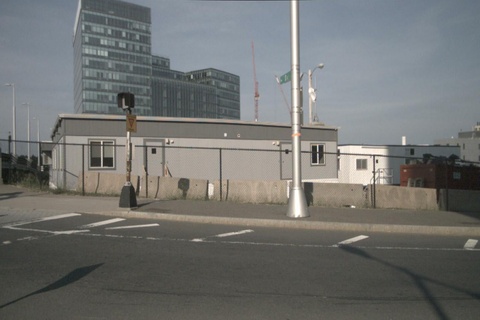}
        \end{minipage}
        \begin{minipage}[b]{.15\linewidth}
            \centering
            \includegraphics[height=1.5cm,width=3cm]{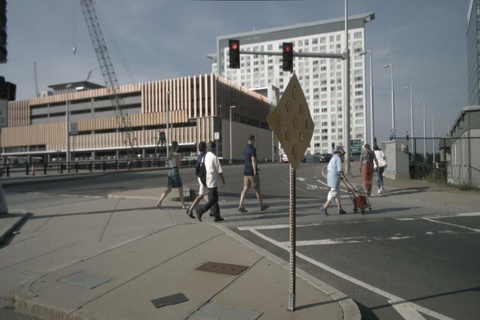}
        \end{minipage}}
    \qquad    
    \subfigure[Depth Map]
    {
        \begin{minipage}[b]{.15\linewidth}
            \centering
            \includegraphics[height=1.5cm,width=3cm]{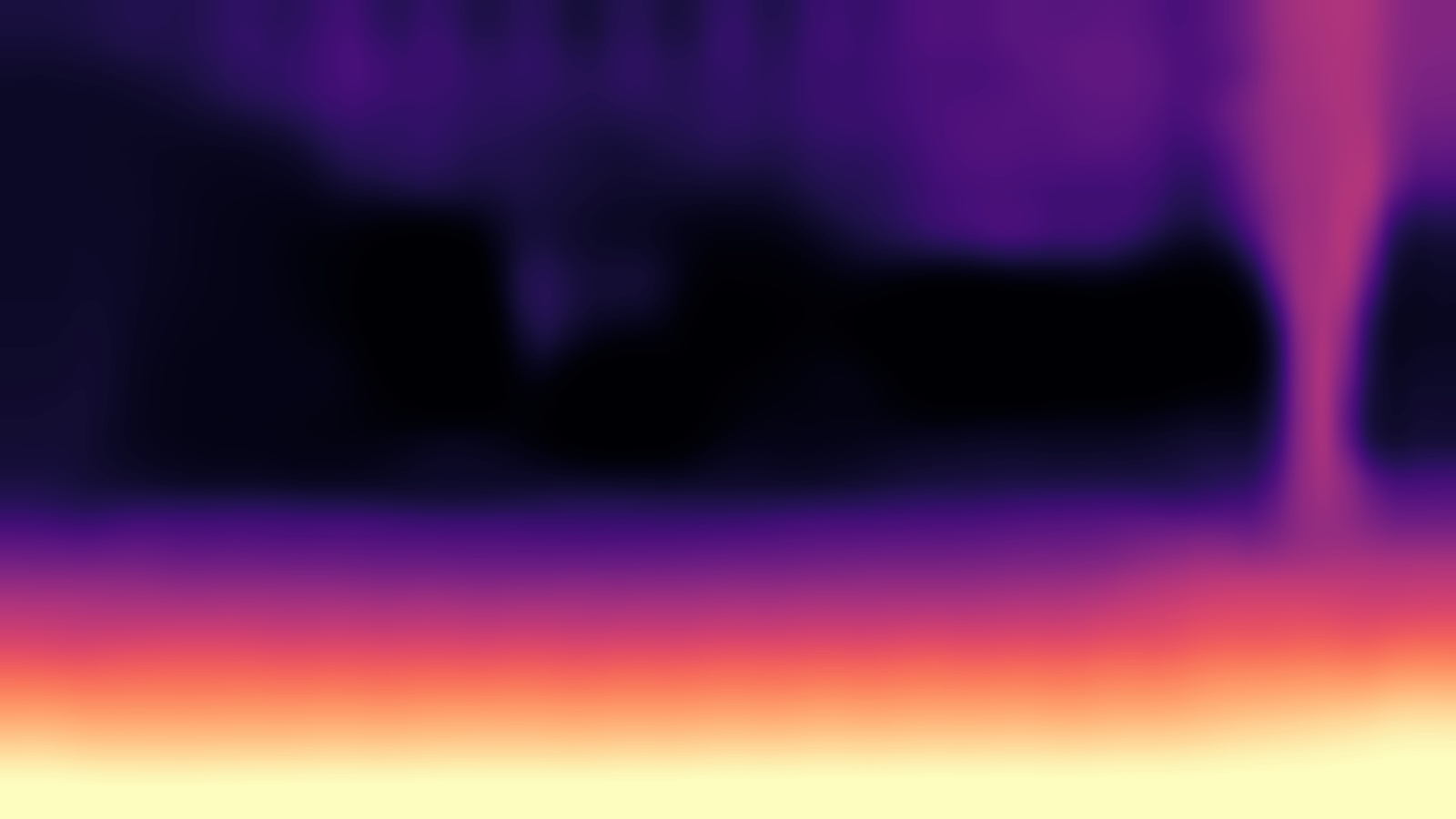}
        \end{minipage}
        \begin{minipage}[b]{.15\linewidth}
            \centering
            \includegraphics[height=1.5cm,width=3cm]{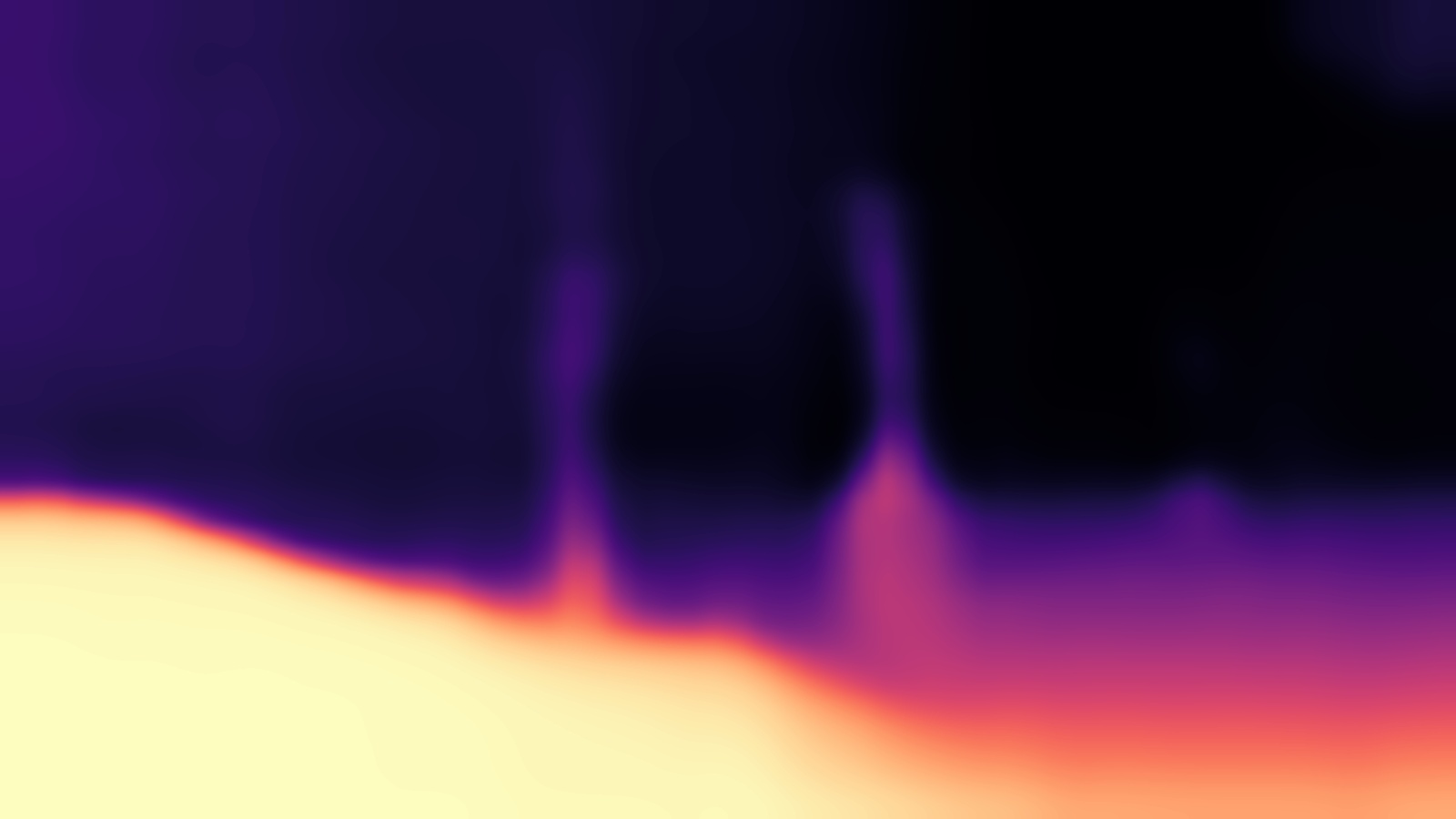}
        \end{minipage}
        \begin{minipage}[b]{.15\linewidth}
            \centering
            \includegraphics[height=1.5cm,width=3cm]{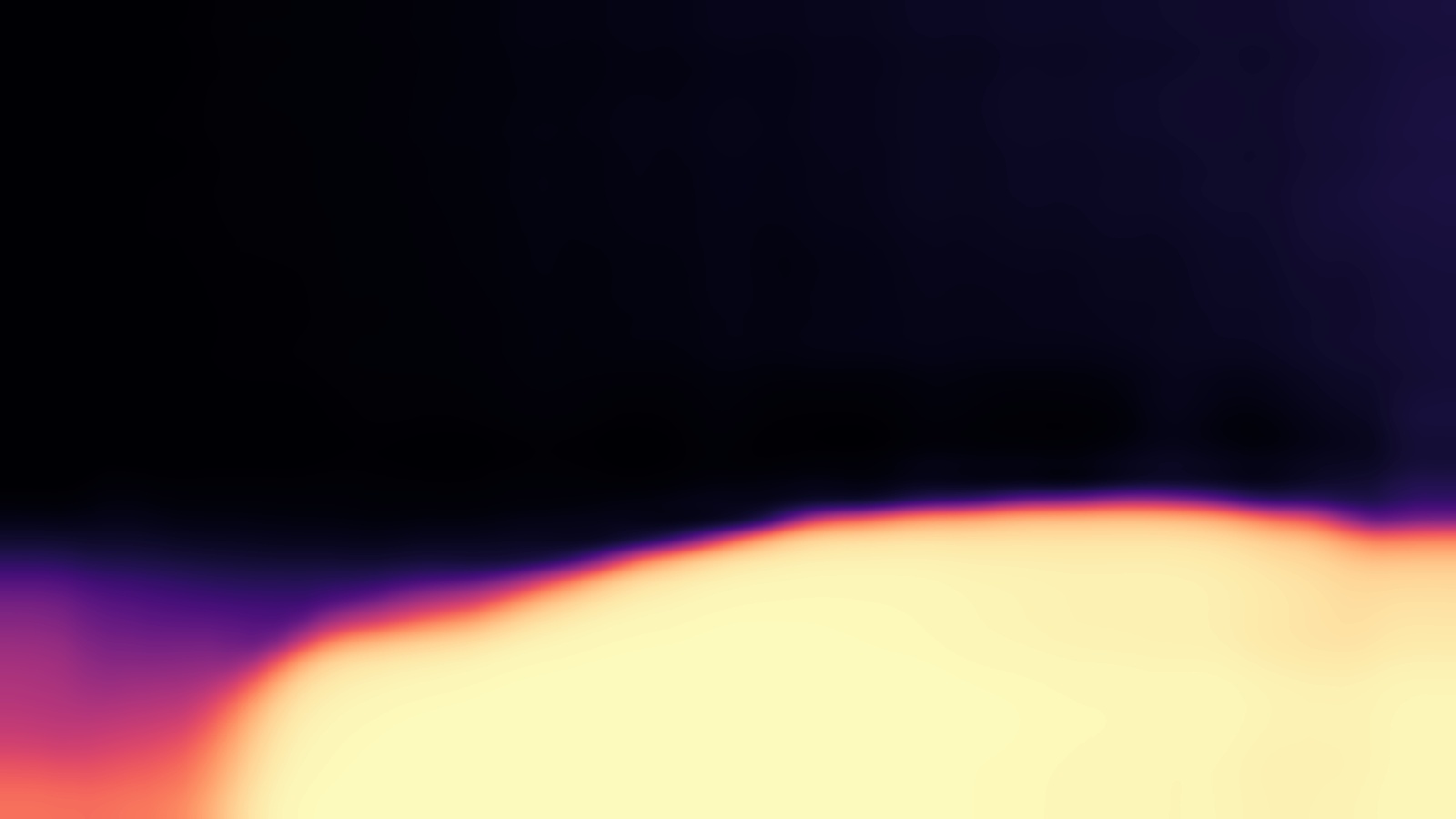}
        \end{minipage}
        \begin{minipage}[b]{.15\linewidth}
            \centering
            \includegraphics[height=1.5cm,width=3cm]{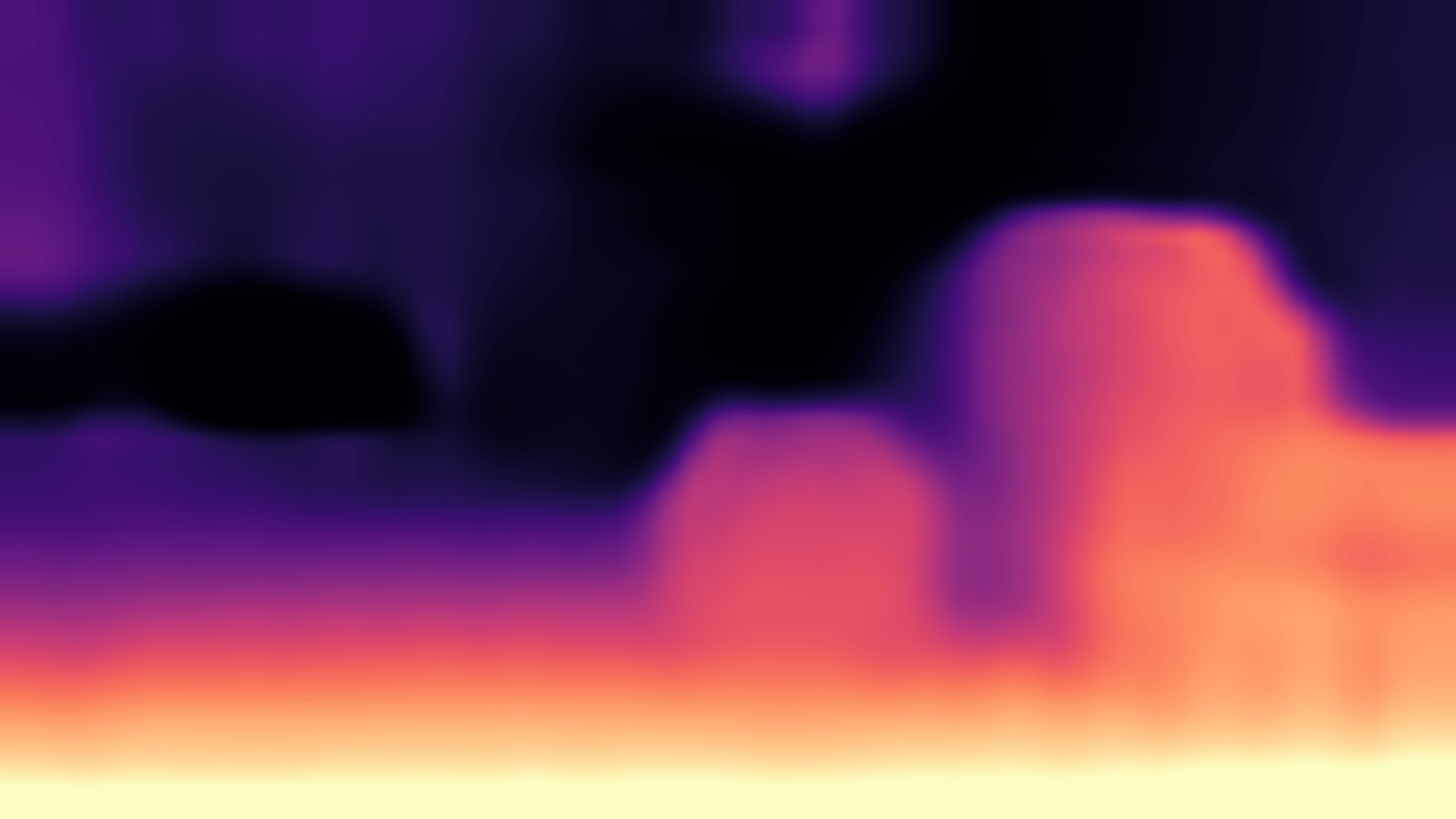}
        \end{minipage}
        \begin{minipage}[b]{.15\linewidth}
            \centering
            \includegraphics[height=1.5cm,width=3cm]{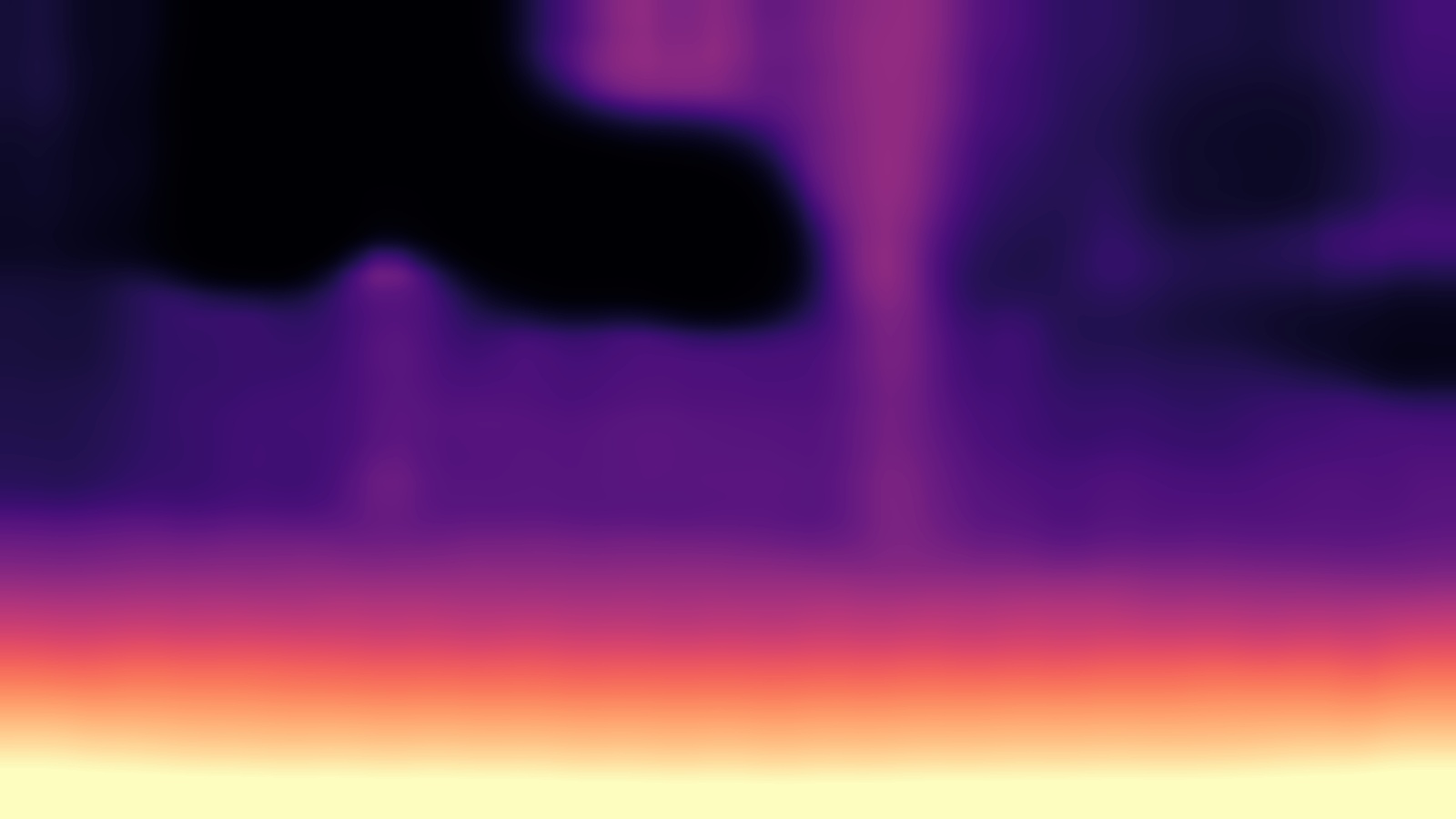}
        \end{minipage}
        \begin{minipage}[b]{.15\linewidth}
            \centering
            \includegraphics[height=1.5cm,width=3cm]{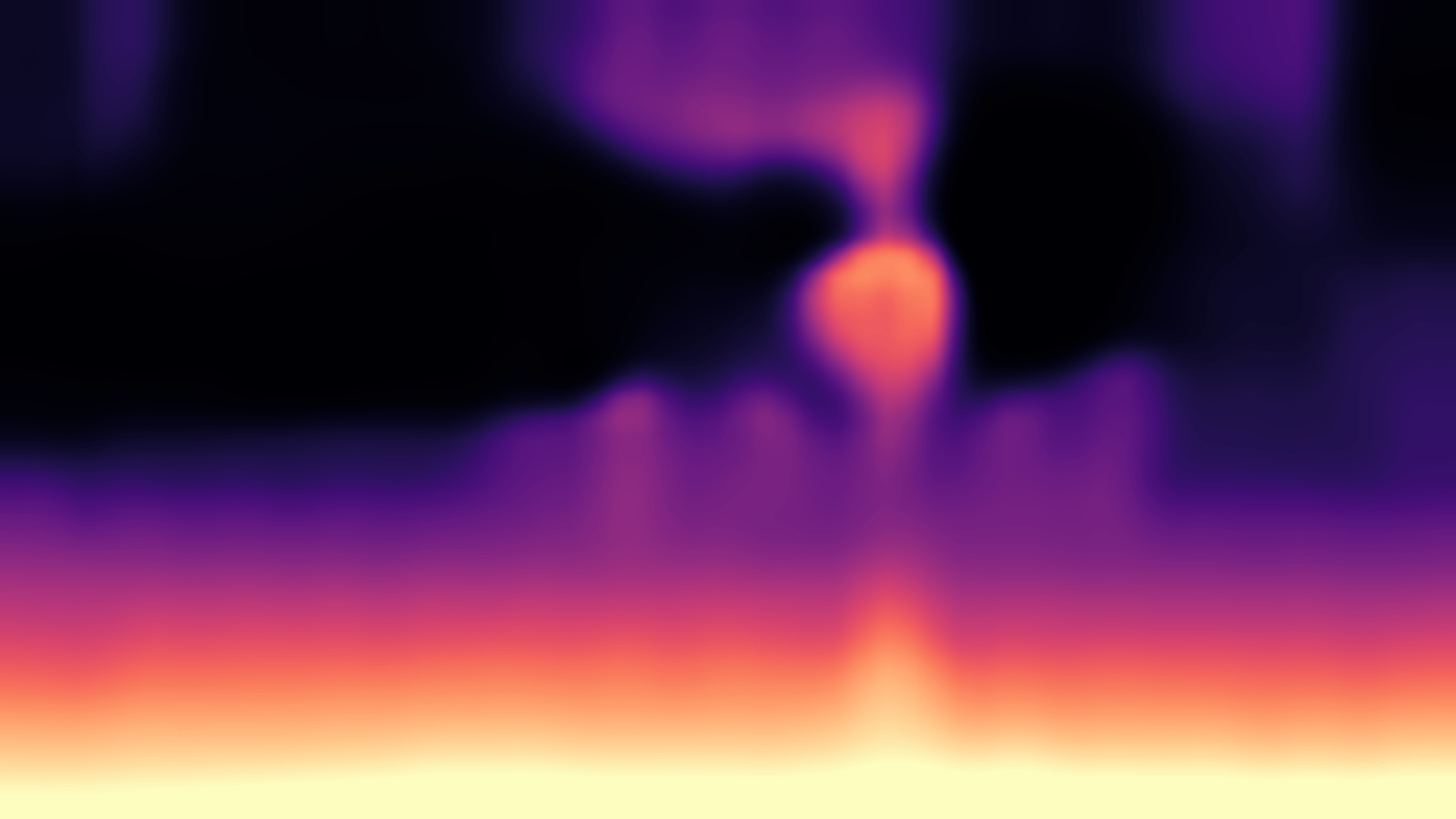}
        \end{minipage}}
    \caption{Qualitative results on nuScenes\cite{nuscenes} datasets.}
    \label{fig5}
\end{figure*}

\subsection{Metrics}
    Self-supervised depth estimation models are typically evaluated using various metrics to assess the quality and accuracy of depth predictions. The following are some used evaluation metrics for our work:

    $$Abs Rel : \frac{1}{N} \sum_{i=1}^{N} \frac{\lvert d_i - \hat{d}_i \rvert}{d_i}$$
    
    $$Sq Rel : \frac{1}{N} \sum_{i=1}^{N} \left(\frac{d_i - \hat{d}_i}{d_i}\right)^2
    $$
    $$RMSE : \sqrt{\frac{1}{N} \sum_{i=1}^{N} (d_i - \hat{d}_i)^2}
    $$
    $$RMSE log : \sqrt{\frac{1}{N} \sum_{i=1}^{N} \left( \log d_i - \log\hat{d}_i \right)^2}
    $$
    \newline
    where $N$ indicates the total number of pixels in a depth image $D$. $d_i$ and $\hat{d}_i$ respectively refer to groundtruth and predicted depth in $i^{th}$ pixel. During the evaluation process, $\frac{median(\hat{D})}{median(D)}$ is typically utilized as  a factor to align the scale.

\begin{figure}[h]
    \centering
    \subfigure[Original Image]
    {
        \begin{minipage}[b]{.49\linewidth}
            \centering
            \includegraphics[height=2cm,width=4.2cm]{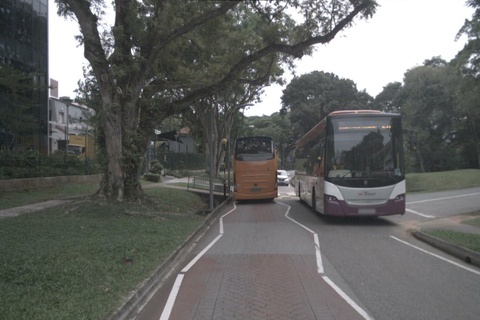}
        \end{minipage}
        \begin{minipage}[b]{.49\linewidth}
            \centering
            \includegraphics[height=2cm,width=4.2cm]{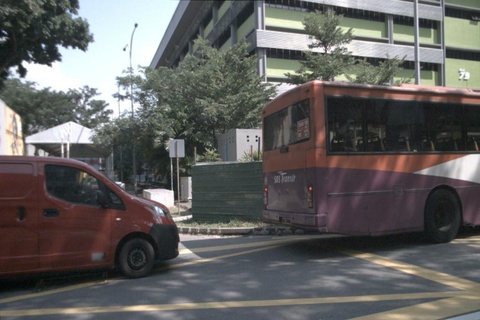}
        \end{minipage}}
    \qquad    
    \subfigure[SurroundDepth]
    {
     	\begin{minipage}[b]{.49\linewidth}
            \centering
            \includegraphics[height=2cm,width=4.2cm]{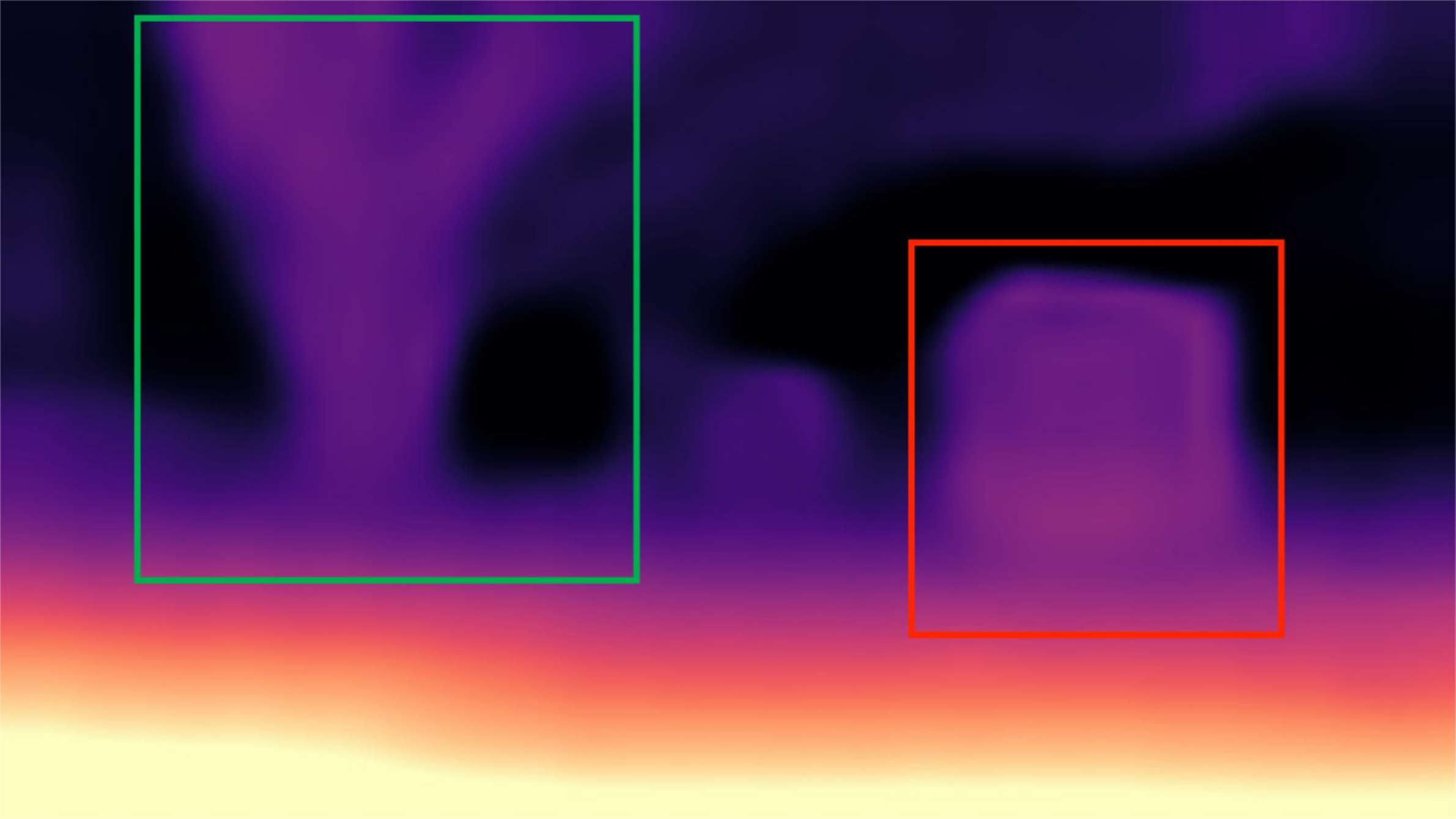}
        \end{minipage}
        \begin{minipage}[b]{.49\linewidth}
            \centering
            \includegraphics[height=2cm,width=4.2cm]{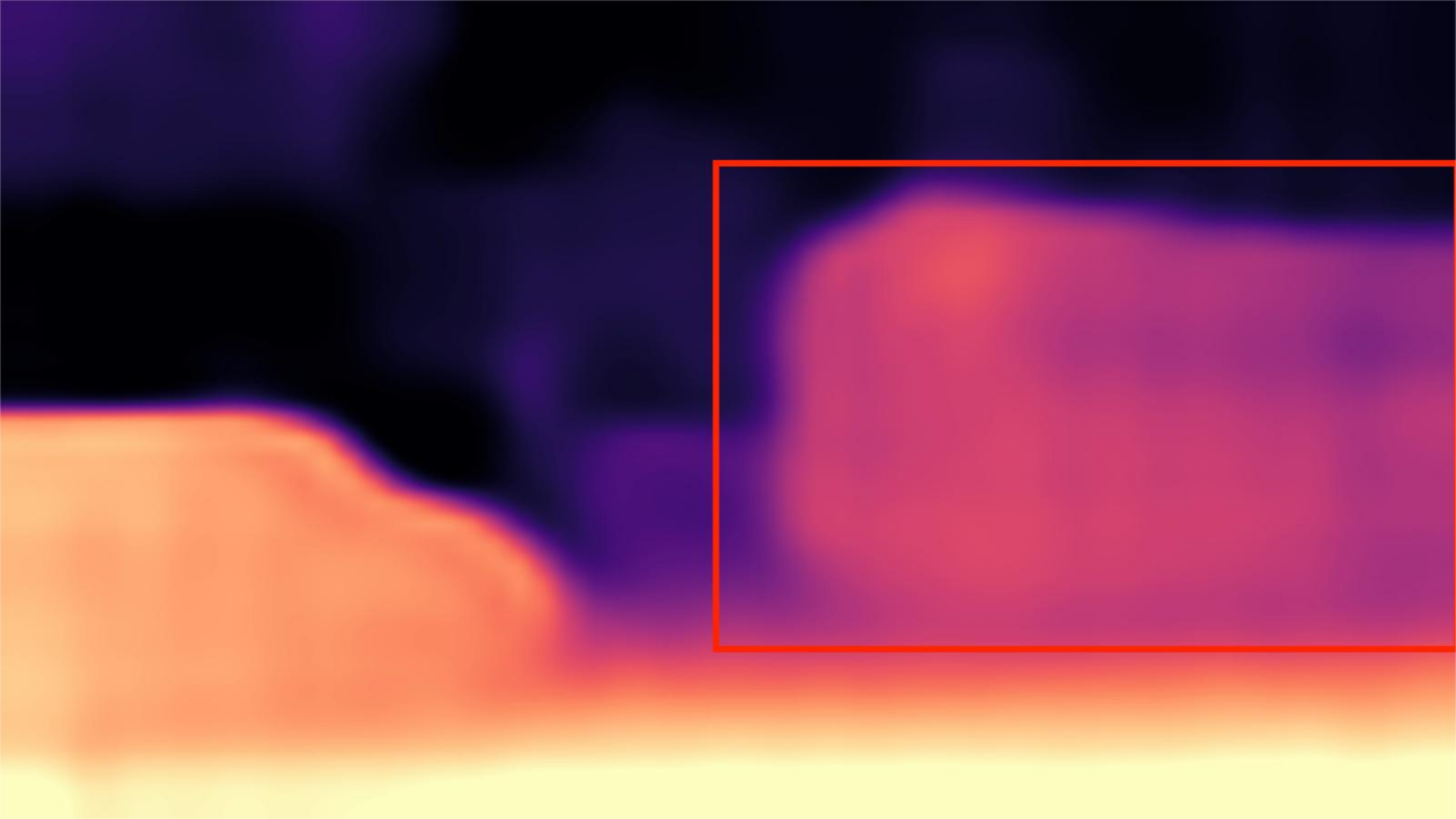}
        \end{minipage}}
    
    \subfigure[BEVScope]
    {
    \begin{minipage}[b]{.49\linewidth}
        \centering
        \includegraphics[height=2cm,width=4.2cm]{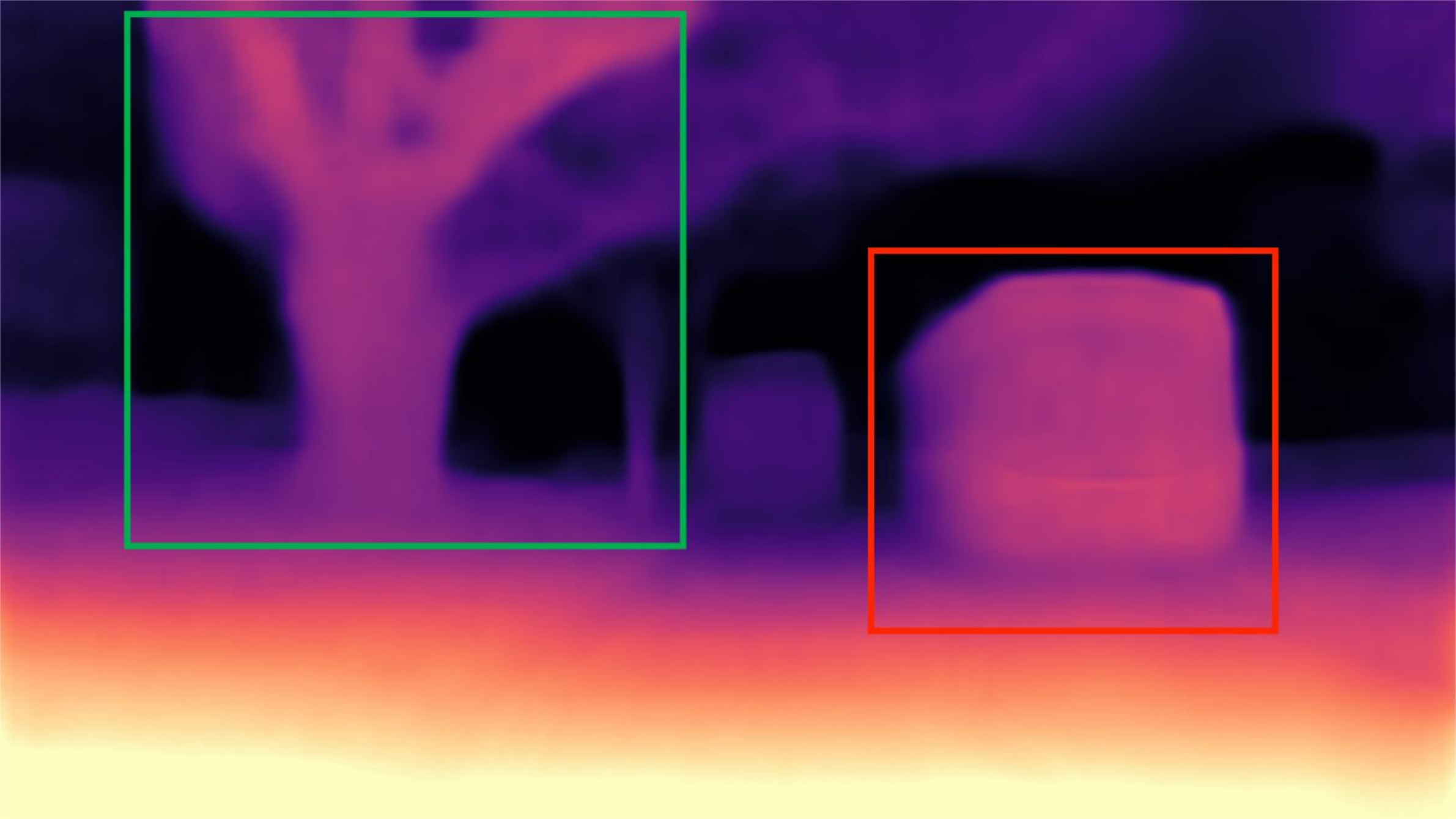}
    \end{minipage}
    \begin{minipage}[b]{.49\linewidth}
        \centering
        \includegraphics[height=2cm,width=4.2cm]{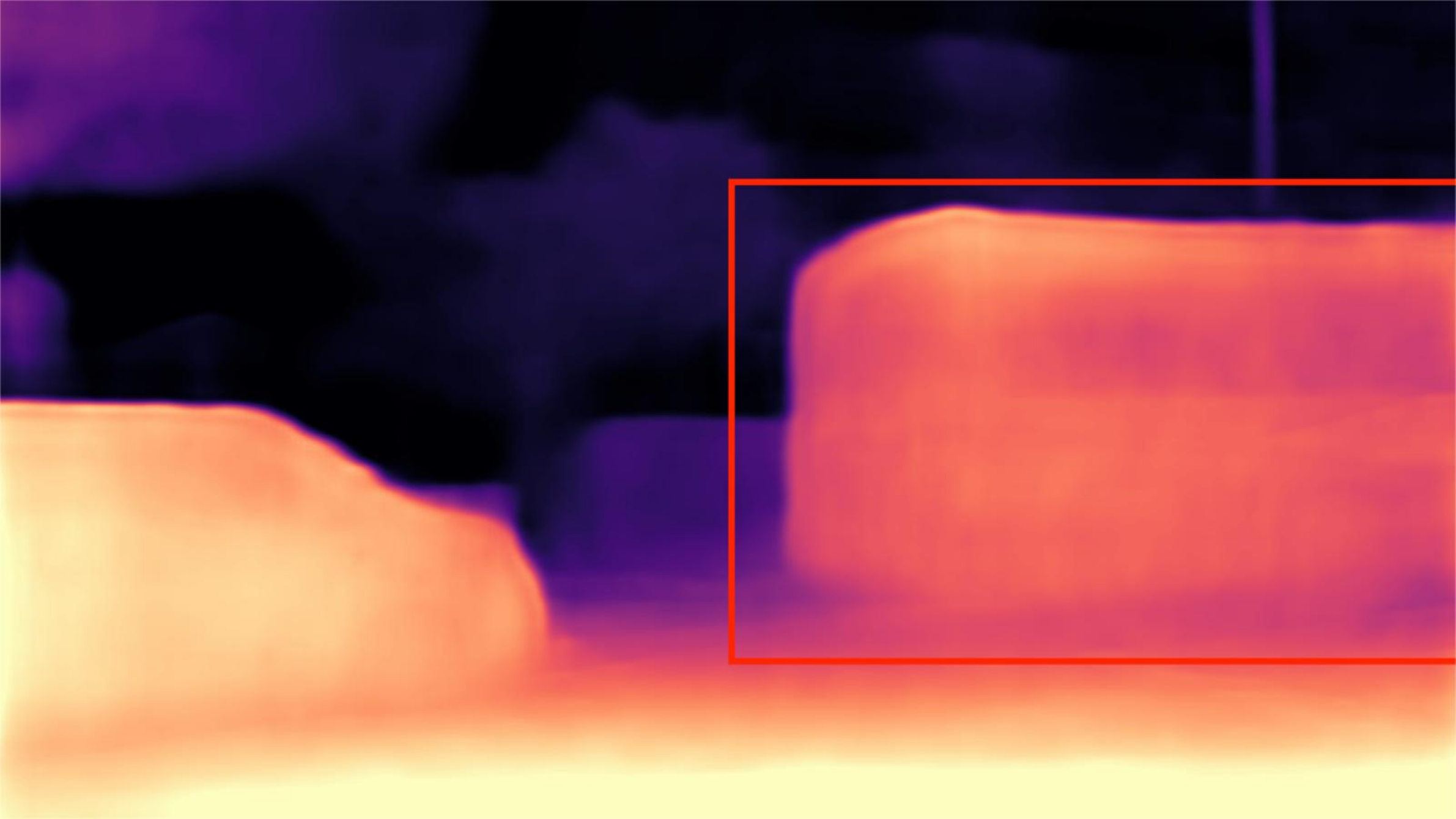}
    \end{minipage}}
    \caption{Comparision of visualization result between Surrounddepth\cite{Surrounddepth}, and the proposed BEVScope.}
    \label{fig6}
\end{figure}

    \subsection{Visualization and Comparison}
    
\begin{figure}[h]
    \centering
    \subfigure[Original Image]
    {
        \begin{minipage}[b]{.49\linewidth}
            \centering
            \includegraphics[height=2.2cm,width=4.2cm]{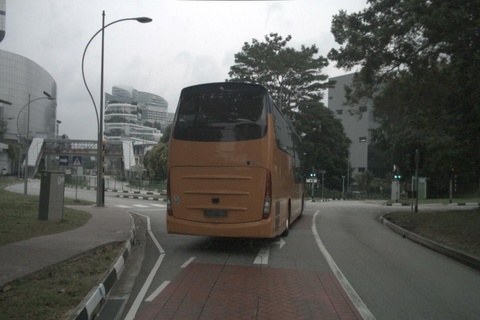}
        \end{minipage}
        \begin{minipage}[b]{.49\linewidth}
            \centering
            \includegraphics[height=2.2cm,width=4.2cm]{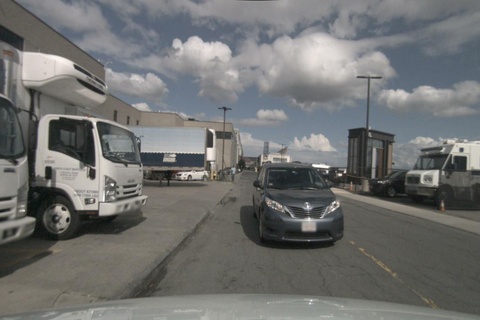}
        \end{minipage}}
    \qquad    
    \subfigure[BEVScope(w/o adaptive mask)]
    {
     	\begin{minipage}[b]{.49\linewidth}
            \centering
            \includegraphics[height=2.2cm,width=4.2cm]{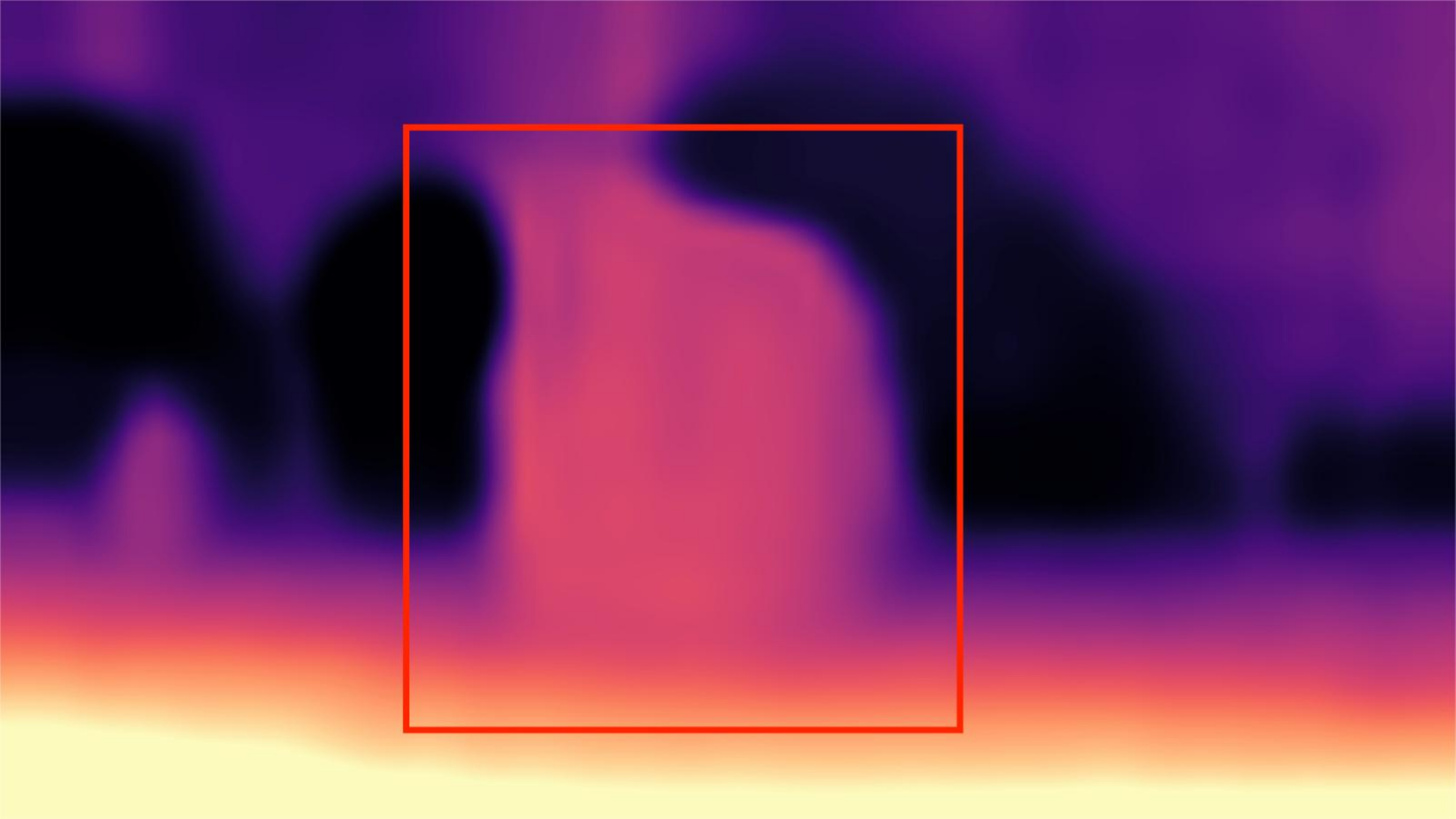}
        \end{minipage}
        \begin{minipage}[b]{.49\linewidth}
            \centering
            \includegraphics[height=2.2cm,width=4.2cm]{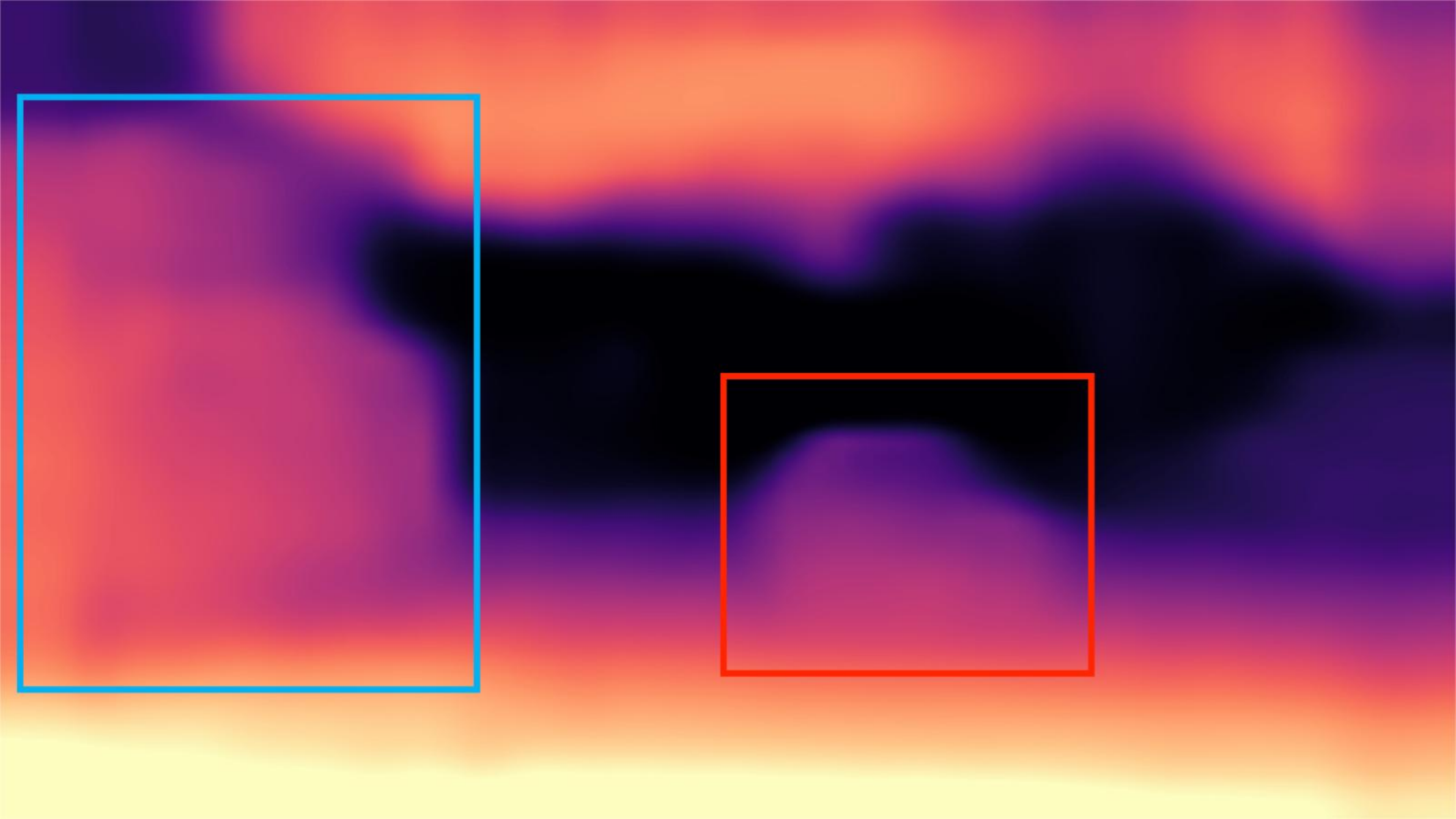}
        \end{minipage}}
    
    \subfigure[BEVScope(w adaptive mask)]
    {
    \begin{minipage}[b]{.49\linewidth}
        \centering
        \includegraphics[height=2.2cm,width=4.2cm]{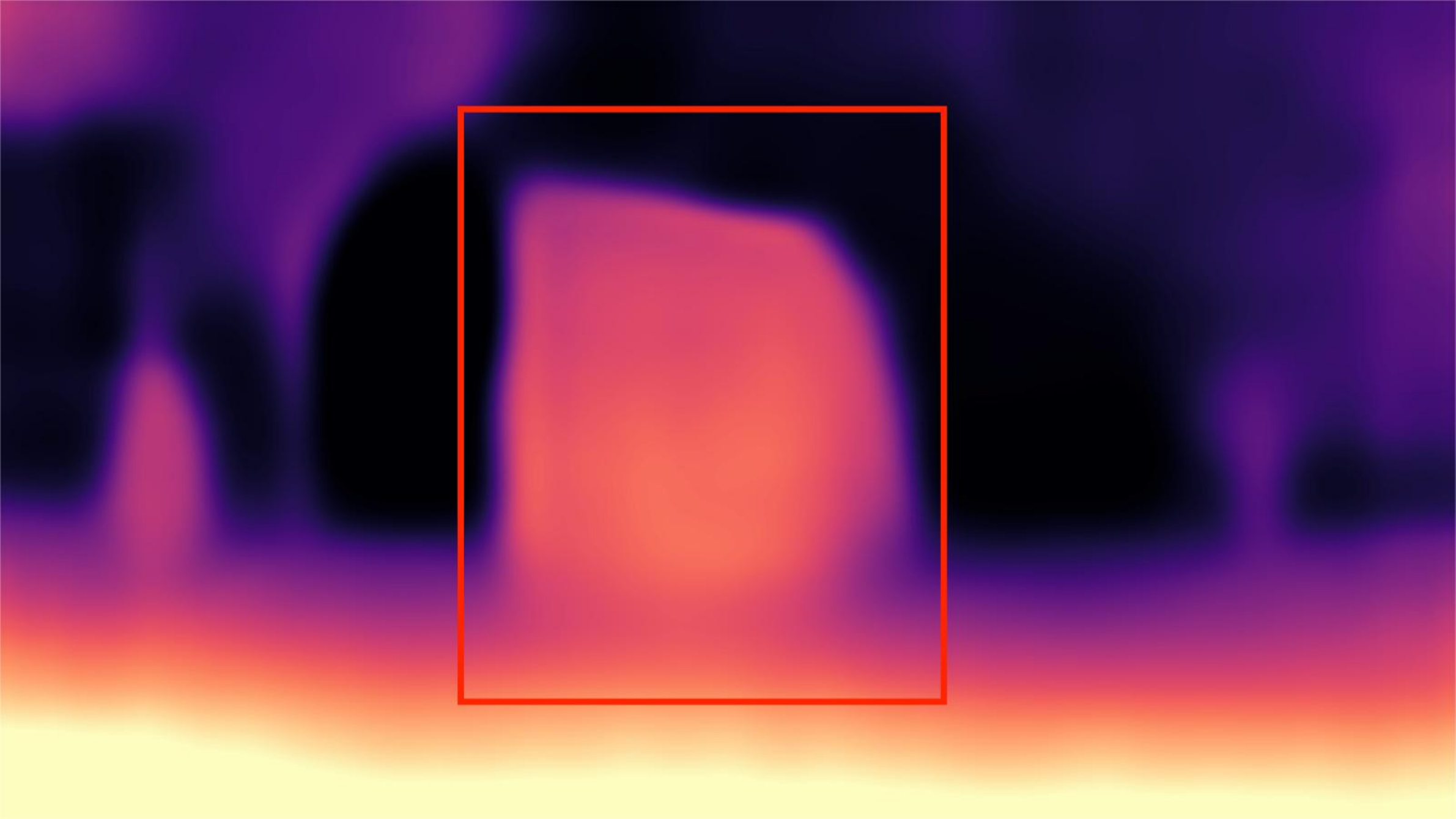}
    \end{minipage}
    \begin{minipage}[b]{.49\linewidth}
        \centering
        \includegraphics[height=2.2cm,width=4.2cm]{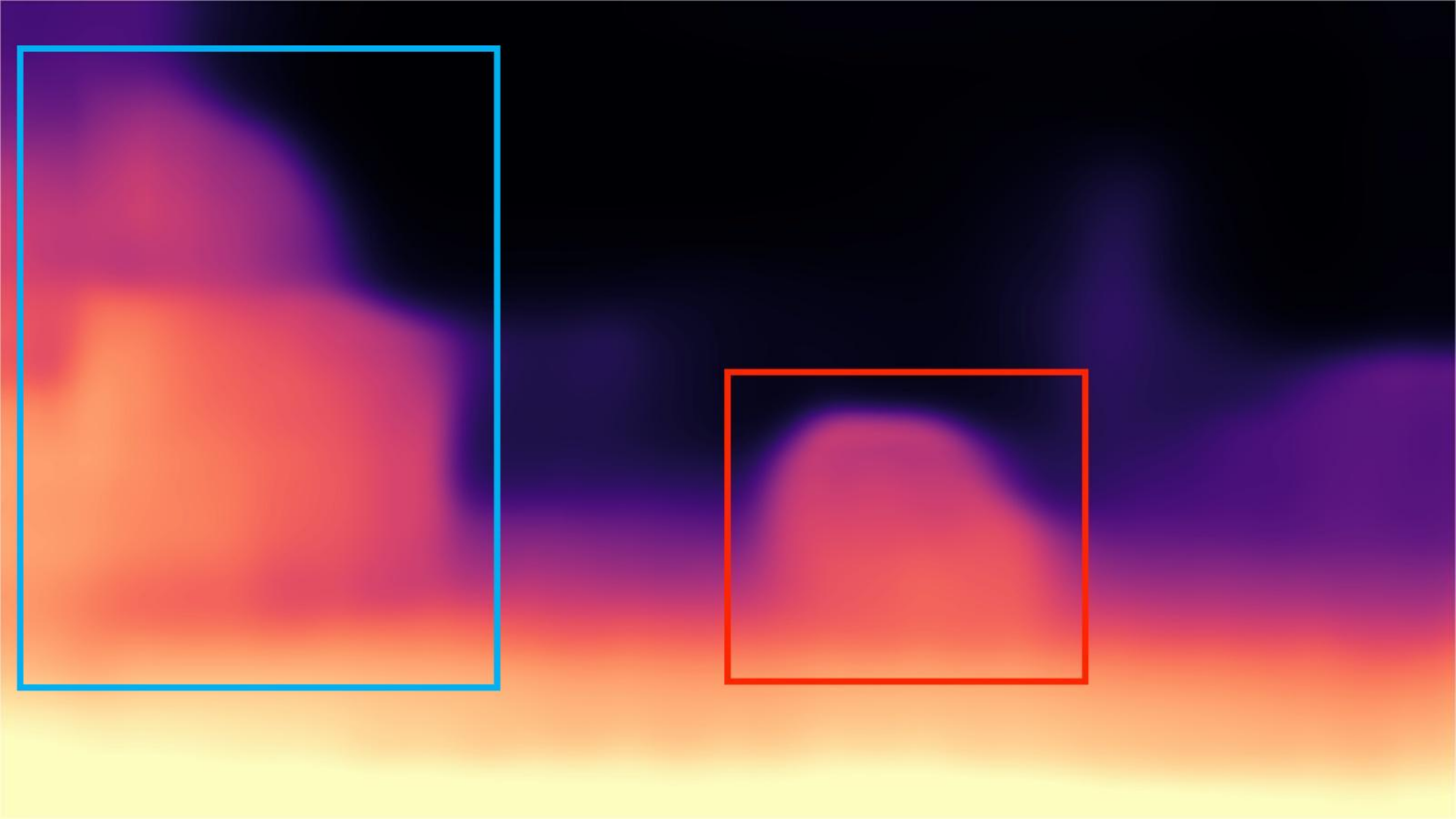}
    \end{minipage}}
    \caption{Comparision between with the adaptive mask and without adaptive mask.}
    \label{fig7}
\end{figure}

    The Figure [\ref{fig5}] is qualitative results on nuScenes\cite{nuscenes} dataset.  As depicted in Figure [\ref{fig6}], we present a qualitative visualization comparison between our model and SurroundDepth. Leveraging robust geometric cues in Bird's Eye View (BEV) features and effective consistency constraints, our model demonstrates remarkable accuracy in generating depth maps. Additionally, we evaluate the prediction outcomes of BEVScope with and without the adaptive mask, as illustrated in Figure [\ref{fig7}]. The inclusion of the adaptive mask results in more precise depth estimation, particularly for rapidly moving vehicles.

\end{document}